\documentclass[lettersize,journal]{IEEEtran}
\usepackage{amsmath,amsfonts}
\usepackage{algorithmic}
\usepackage{algorithm}
\usepackage{array}
\usepackage[caption=false,font=normalsize,labelfont=sf,textfont=sf]{subfig}
\usepackage{textcomp}
\usepackage{stfloats}
\usepackage{url}
\usepackage{verbatim}
\usepackage{graphicx}
\usepackage{multirow}
\usepackage{cite}
\usepackage{colortbl}  
\usepackage{xcolor}
\usepackage{adjustbox}
\usepackage{tabularx}
\usepackage{pifont} 
\usepackage{booktabs}
\usepackage[normalem]{ulem}

\begin{document}

\title{FMC-DETR: Frequency-Decoupled MultiDomain Coordination for Aerial-View Object Detection}

\author{Ben Liang,~\IEEEmembership{Graduate Student Member,~IEEE}, Hongguang Wei, Yuan Liu,~\IEEEmembership{Member,~IEEE}, Bingwen Qiu, Yihong Wang, Xiubao Sui, and Qian Chen
\thanks{This work was supported in part by the National Natural Science Foundation
of China under Grant 62427818, Grant U25A20502, and Grant 62301253; in part by the Fundamental Research Funds for the Central Universities under Grant 30925010511; and in part by the Postgraduate Research \& Practice Innovation Program of Jiangsu Province under Grant 26CXJH1111. (Ben Liang and Hongguang Wei contributed equally to this work.) (Corresponding author: Yuan Liu and Xiubao Sui.)}

\thanks{Ben Liang, Hongguang Wei, Yuan Liu, Bingwen Qiu, Yihong Wang, Xiubao Sui, and Qian Chen are with the School of Electronic and Optical Engineering, Nanjing University of Science and Technology, Nanjing 210094, China (e-mail: benliang@njust.edu.cn; weihg@njust.edu.cn; lyuan90\_eo@njust.edu.cn; qbw0315@njust.edu.cn; h@njust.edu.cn; sxb@njust.edu.cn; chenqian@njust.edu.cn).}
\thanks{Xiubao Sui is also with the State Key Laboratory of Extreme Environment Optoelectronic Dynamic Measurement Technology and Instrument, Nanjing 210094, China.}
}

\markboth{Journal of \LaTeX\ Class Files,~Vol.~14, No.~8, August~2021}%
{Shell \MakeLowercase{\textit{et al.}}: A Sample Article Using IEEEtran.cls for IEEE Journals}


\maketitle

\begin{abstract}
Remote sensing object detection is a critical technology for real-world applications such as natural resource monitoring, traffic management, and UAV-based rescue. Detecting tiny objects in high-resolution aerial imagery remains challenging due to weak visual cues and insufficient global context modeling in complex scenes. Existing methods often suffer from delayed contextual interaction and limited nonlinear reasoning, which restrict their ability to effectively refine shallow representations and ultimately lead to suboptimal performance. To address these challenges, we propose FMC-DETR, a frequency-decoupled fusion framework for aerial-view object detection. First, we propose the Wavelet Kolmogorov-Arnold Transformer (WeKat) backbone, which employs cascaded wavelet transforms to enhance global low-frequency structure perception in shallow features while preserving fine-grained details, and further leverages Kolmogorov-Arnold networks for adaptive nonlinear modeling of multi-scale dependencies. Second, we introduce the Multi-Domain Feature Coordination (MDFC) module, which refines cross-scale fused representations through partial-channel spatial, spectral, and structural coordination, thereby strengthening small-object-related feature responses in cluttered scenes. Finally, we design the Compact Partial Fusion (CPF) module, which performs compact multi-branch aggregation with progressive partial refinement to improve feature diversity and multi-scale interaction while preserving stable information flow and reducing redundant perturbation. 
Extensive experiments across multiple remote sensing benchmarks demonstrate that FMC-DETR achieves competitive performance and consistently improves over the baseline detector. Code is available at \url{https://github.com/bloomingvision/FMC-DETR.}
\end{abstract}


\section{Introduction}
Object detection, a cornerstone task in computer vision, has demonstrated profound utility in a multitude of industrial applications, including autonomous driving \cite{lin2024rcbevdet,zhang2025sfnet}, urban surveillance \cite{pang2021facet,he2024wtapnet,pang2024lrta,chen2024dtssnet,ge2025sga}, and remote sensing \cite{liu2025hyperspectral,wei2024spatio,wei2025multi,wei2026foreground}. Propelled by the rapid advances in deep learning and imaging technologies \cite{qian2025ensemble,chen2024deep,liu2024scene}, particularly Convolutional Neural Networks (CNNs), modern detection frameworks have achieved remarkable success on natural images at relatively low resolution. However, these detectors often suffer from significant performance degradation on high-resolution remote sensing imagery, where objects are typically small, weakly featured, randomly distributed, and frequently occluded \cite{xiao2025fbrt,liang2026multi}.

\begin{figure}[!t]
\centering
\includegraphics[width=.6\columnwidth]{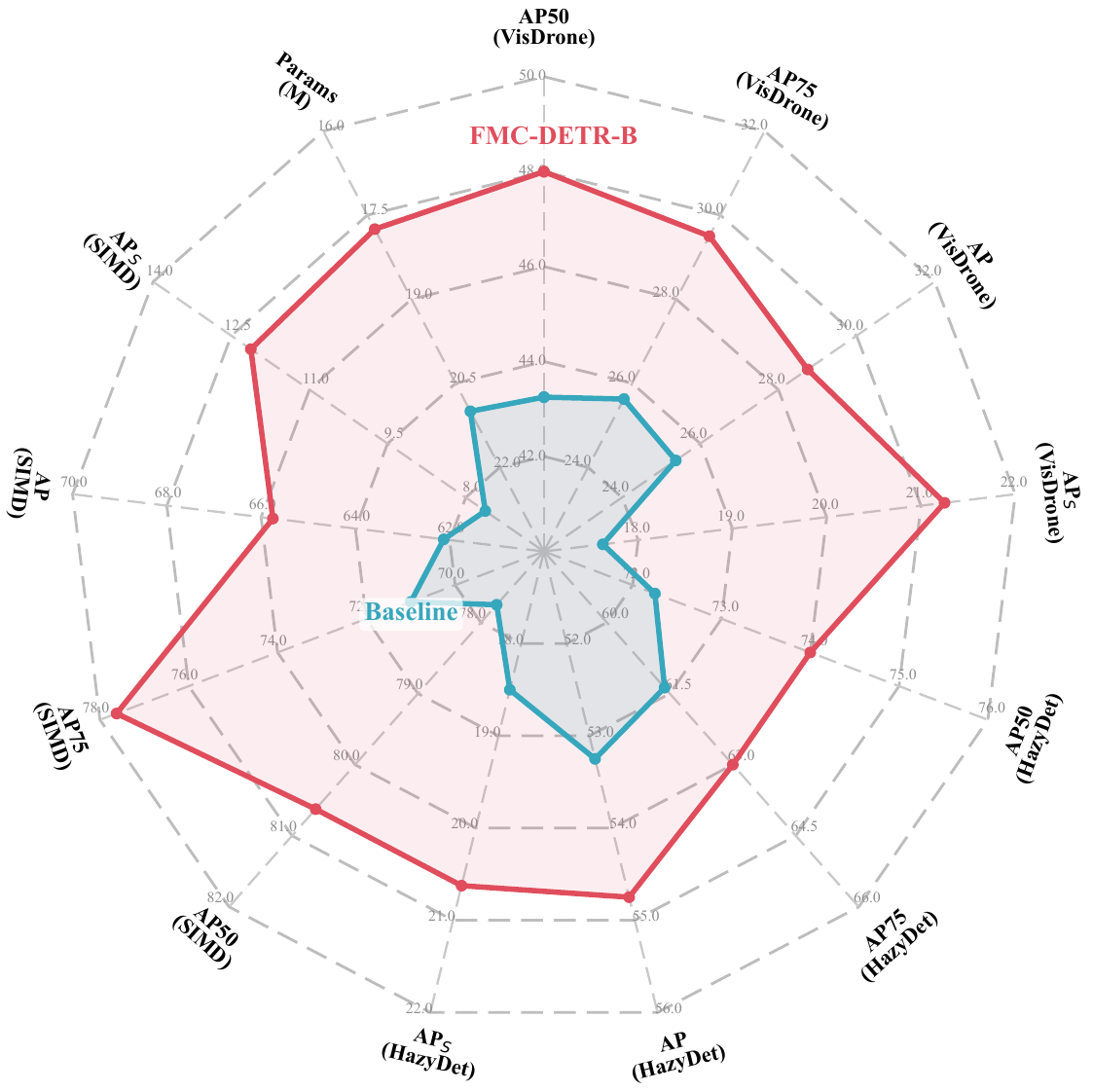} 
\caption{Performance comparison between FMC-DETR-B and the Baseline \cite{zhao2024detrs} on the VisDrone, HazyDet, and SIMD datasets in terms of AP$_{50}$, AP$_{75}$, AP, AP$_{S}$, and model parameters. The outer regions indicate better performance.}
\label{fig-radar}
\end{figure}

To mitigate the limited appearance information of small objects, contextual modeling has become an important direction for remote sensing object detection (RSOD) \cite{deng2020global,min2023yolo,liu2024hyperspectral}. CNNs are effective at preserving local structures, but their inherently local inductive bias limits the direct modeling of long-range dependencies. Transformers alleviate this limitation through self-attention \cite{vaswani2017attention,dosovitskiy2020image}; however, the quadratic complexity with respect to the number of tokens makes global interaction on high-resolution shallow features computationally expensive. Consequently, most hybrid architectures retain convolution-dominated processing in shallow stages and introduce stronger global interaction only after substantial spatial downsampling. Such a design, however, overlooks an important characteristic of aerial small-object representation: different stages require not only different receptive fields, but also different transformation strengths. Shallow features retain the weak boundaries and fine structures of tiny objects, and excessive transformation may easily disturb these already scarce cues; meanwhile, their locally dominated representations still require broader structural perception to distinguish targets from complex backgrounds. In deeper stages, where spatial redundancy is reduced and semantic abstraction becomes dominant, stronger long-range interaction and more flexible nonlinear transformation are desirable. A more appropriate backbone should explicitly separate information preservation from strong feature transformation, dynamically regulate their interaction, and concentrate expressive nonlinear modeling on the feature subsets and stages where it is most beneficial.

Moreover, multi-scale feature fusion has become a common strategy for aerial small-object detection. However, features from different stages exhibit substantial discrepancies in spatial details, semantic abstraction, and background responses. Shallow features preserve valuable fine-grained structures for tiny objects, but are also accompanied by abundant background interference, whereas deeper features provide stronger semantic information with weakened spatial details. Direct cross-scale aggregation therefore cannot fully exploit their complementarity and may propagate irrelevant responses into the fused representation. Furthermore, conventional fusion modules typically apply uniform transformation to all aggregated features, which may introduce redundant computation and disturb already informative responses. These observations motivate a more coordinated fusion strategy that selectively transfers shallow details, reconciles heterogeneous multi-scale information, and progressively refines informative features while maintaining stable information propagation.

To address these limitations, we propose FMC-DETR, a frequency-aware collaborative framework that follows a unified principle of selective representation coordination across stages, scales, and feature subsets. First, the Wavelet Kolmogorov--Arnold Transformer (WeKat) addresses stage-wise representation mismatch through a heterogeneous split-gating (HSG) backbone. In shallow high-resolution stages, HSG-WAVE performs wavelet-based structure--detail decoupling within the computation stream, strengthening low-frequency structural perception while retaining fine-grained information through low-perturbation pathways. In deeper stages, HSG-AKAT selectively introduces asymmetric self-attention and Group KAN into the computation stream to enhance global semantic interaction and adaptive nonlinear transformation. Second, Multi-Domain Feature Coordination (MDFC) converts the shallow $S2$ output into an amplitude-guided detail prior before scale alignment and then refines the fused feature through spatial, spectral, and frequency-guided edge responses. This design enables explicit multi-domain coordination during cross-scale fusion, allowing complementary spatial, spectral, and structural cues to jointly enhance the aggregated representation. Finally, Compact Partial Fusion (CPF) adopts a structured multi-branch fusion paradigm that combines stable feature preservation with progressive partial refinement, thereby promoting stronger cross-scale feature complementarity and more expressive aggregation. Consequently, WeKat, MDFC, and CPF establish a progressive collaborative pipeline that performs stage-aware representation adaptation, cross-scale multi-domain coordination, and compact selective refinement in sequence. Extensive experiments on publicly available remote sensing datasets, including VisDrone \cite{cao2021visdrone}, HazyDet \cite{feng2026hazydet}, and SIMD \cite{haroon2020multisized}, demonstrate the effectiveness of FMC-DETR over the baseline detector. As shown in Fig.~\ref{fig-radar}, FMC-DETR-B achieves AP scores of 29.4\%, 55.0\%, and 66.6\% on VisDrone, HazyDet, and SIMD, respectively. The main contributions of this work are summarized as follows:
\begin{itemize}

\item We propose WeKat, a stage-aware backbone built on a unified heterogeneous split-gating mechanism. HSG-WAVE coordinates structural and fine-grained information in shallow high-resolution stages, whereas HSG-AKAT selectively strengthens global semantic interaction and adaptive nonlinear transformation in deeper stages. This stage-specific heterogeneous computation enables WeKat to adapt feature transformation to the distinct representation requirements of different backbone stages.

\item We propose MDFC to explicitly coordinate heterogeneous information across scales. It first converts shallow high-resolution features into an amplitude-guided detail prior, and then jointly refines the fused representation from spatial, spectral, and edge-aware structural perspectives, thereby enhancing tiny-object feature saliency and boundary-aware representation.

\item We propose CPF, a compact aggregation module that combines direct feature interaction, stable preservation pathways, and progressive partial-channel refinement. By selectively strengthening only a subset of features while retaining low-perturbation information flow, CPF improves feature complementarity and aggregation efficiency without repeatedly transforming all channels.
\end{itemize}
\section{Related Work}
\subsection{CNN--Transformer Object Detectors}
CNNs and Transformers constitute two major paradigms for modern object detection. CNN-based detectors, represented by the YOLO series \cite{yolov8_ultralytics,wang2024yolov9,wang2024yolov10,yolo11_ultralytics,tian2025yolov12,lei2025yolov13}, achieve a favorable balance between local representation capability and inference efficiency. Transformers \cite{dosovitskiy2020image}, in contrast, introduce self-attention to capture long-range dependencies and global contextual interactions. Recent DETR-based detectors, including DETR \cite{carion2020end}, Deformable DETR \cite{zhu2020deformable}, RT-DETR \cite{zhao2024detrs}, D-FINE \cite{peng2024d}, and DEIM \cite{huang2025deim}, further combine convolutional feature extraction with attention-based global interaction, establishing an effective hybrid detection paradigm. Existing hybrid architectures mainly distribute convolution and attention according to network depth, while feature transformation within each stage remains relatively homogeneous. WeKat instead adopts a unified HSG-based selective computation paradigm, which preserves stable information while allocating stronger transformation to informative feature components. Under this framework, shallow stages emphasize structure--detail coordination through frequency-aware modeling, whereas deeper stages focus on global semantic interaction and adaptive nonlinear representation.

\subsection{Small Object Detection}
Small-object detection remains challenging due to limited visual cues, complex backgrounds, and information loss during feature extraction. Recent studies have therefore explored contextual enhancement and multi-scale representation learning. GLSAN \cite{deng2020global} integrates global--local fusion into a progressive scale-changing network to improve object localization. CEASC \cite{du2023adaptive} introduces context-enhanced sparse convolution to strengthen global information, while YOLO-DCTI \cite{min2023yolo} employs Contextual Transformer to improve the utilization of contextual cues for small objects. FBRT-YOLO \cite{xiao2025fbrt} further introduces feature complementary mapping to coordinate semantic and spatial information for improved small-object localization. In comparison, our method jointly addresses stage-specific representation, cross-scale feature coordination, and selective feature refinement, forming a unified framework for preserving fine-grained cues while strengthening contextual and semantic modeling for aerial small objects. Recent advances in KAN-based modeling have also shown the potential of adaptive nonlinear transformation for visual representation. KAT \cite{yang2024kolmogorov} introduces Group KAN into Transformer architectures for scalable nonlinear feature transformation, while KSCNet \cite{li2025kscnet} combines KAN-based feature extraction with state-space modeling for small-object detection. These methods primarily exploit KAN as an enhanced nonlinear representation operator. Our WeKat assigns Group KAN a stage-specific role within a heterogeneous backbone, using it to strengthen deep semantic abstraction while shallow stages focus on structure--detail representation.

\subsection{Frequency-domain Fusion}
Frequency-domain features have attracted considerable attention in object detection due to their ability to reveal intricate structural information across diverse scenes. Unlike spatial-domain features typically extracted by conventional convolutions, which capture spatial layout and contextual relationships in an implicit and localized manner, frequency-domain analysis provides a more explicit and global perspective by decomposing intensity variations into distinct spectral components. SWANet \cite{he2023low} uses wavelet transforms to perform multi-scale decomposition of noise and details in images, effectively improving image quality under low light conditions. ELWNet \cite{wang2023elwnet} combines the wavelet transform module with CNNs for feature downsampling, using finite parameters to achieve high-quality multi-level encoded features. SFS-CNet \cite{li2024unleashing} proposes a space-frequency selective convolution, which employs a diversion-perception selection strategy to enhance the diversity and uniqueness of features, thereby improving the performance of SAR target detection. Existing frequency-domain methods mainly employ spectral transforms as within-stage operators for feature enhancement, denoising, downsampling, or selective representation learning. In FMC-DETR, frequency modeling is further organized according to different representation roles across the network. HSG-WAVE exploits wavelet decomposition in shallow high-resolution stages to coordinate structural information and fine-grained details, while MDFC introduces frequency cues into cross-scale fusion for shallow-detail calibration and multi-domain feature coordination. This stage- and fusion-aware frequency modeling establishes a more coherent spectral information flow for aerial small-object detection.

\section{Method}
\begin{figure*}[t]
\centering
\includegraphics[width=\textwidth]{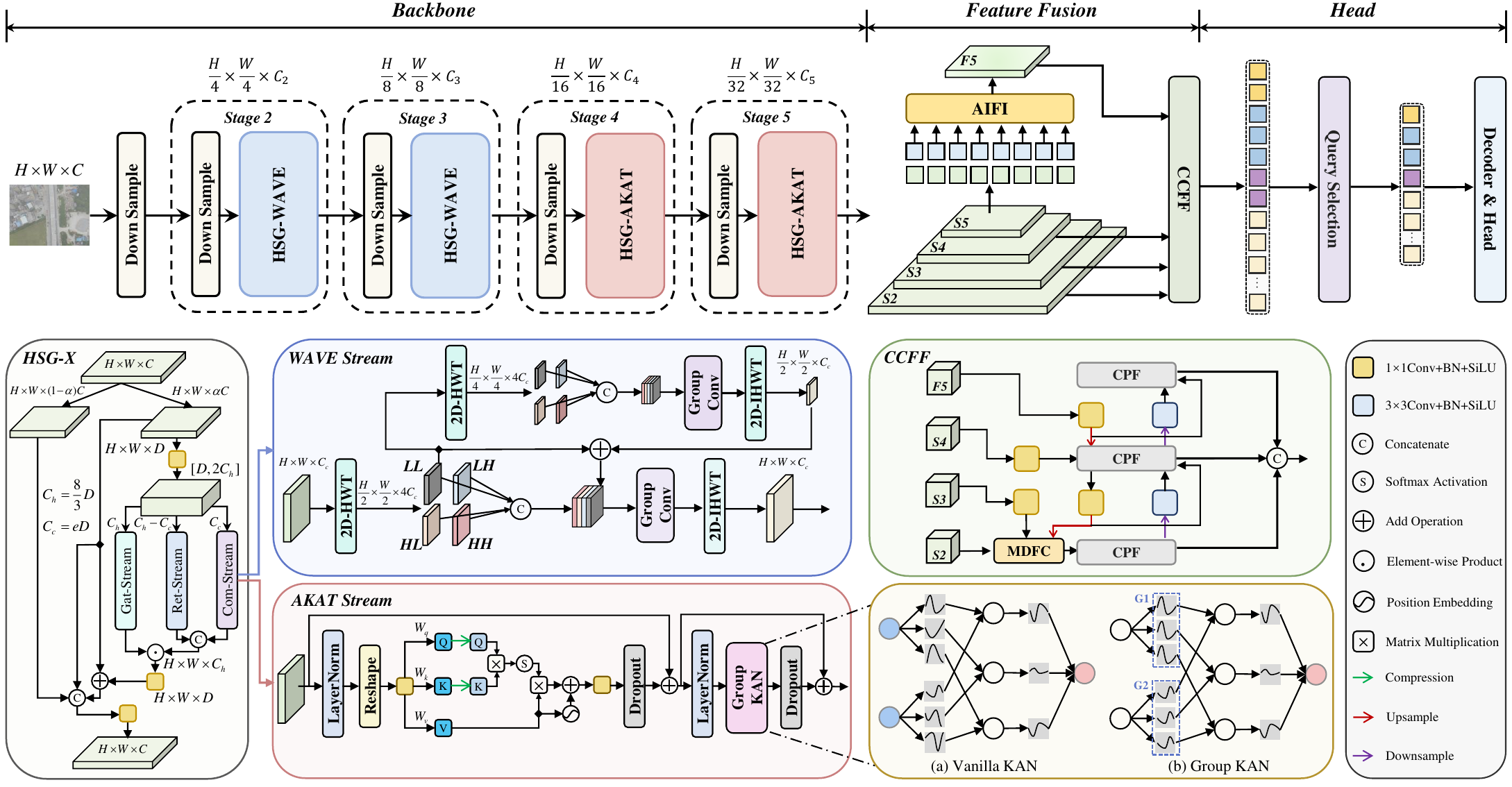} 
\caption{Overview of the proposed FMC-DETR. The backbone consists of multiple stages to extract and model multi-scale features. Among them, HSG-WAVE and HSG-AKAT are applied to capture correlations in shallow and deep features, respectively. The multi-scale features are then fed into the CCFF structure for feature fusion. Finally, the Decoder and Detection Head query the optimized features to produce the final detection results.}
\label{fig1}
\end{figure*}

In this section, we introduce the proposed model, named FMC-DETR. As illustrated in Fig.~\ref{fig1}, FMC-DETR employs a cascaded hierarchical design, consisting of three main stages from a macro perspective: the backbone network extracts feature maps at various scales, the cross-layer feature interaction structure facilitates the fusion of multi-scale contextual information, and the detection head performs both feature classification and bounding box localization.
\subsection{Wavelet Kolmogorov-Arnold Transformer}
In recent years, hybrid CNN-Transformer backbones have shown strong potential for visual representation learning \cite{carion2020end,zhao2024detrs,yang2024ssgcrtn,yang2026decoupled}. Yet a stage-wise mismatch remains: shallow high-resolution features often couple under-exploited low-frequency structures with dominant high-frequency details, causing conventional convolutions to favor local textures over stable structural cues. In contrast, deeper semantic stages require selective rather than uniform transformation, since structural information should be preserved while more informative components benefit from stronger nonlinear modeling. To address this issue, we propose the Wavelet Kolmogorov-Arnold Transformer (WeKat), a hybrid backbone built on a Heterogeneous Split-Gating (HSG) design. As shown in Fig.~\ref{fig1}, HSG decomposes feature propagation into retention, gating, and computation streams to enable low-perturbation information preservation, dynamic feature modulation, and controlled strong transformation, respectively. 

Given a stage feature with channel width $C$, HSG first constructs an internal representation with width $D= \alpha C$, where $\alpha=0.5$ by default. A pointwise projection maps the processing branch from $D$ to $2C_h$ channels, where $C_h=\lfloor\frac{8}{3}D\rfloor$. The projected feature is then partitioned into the Gat-, Ret-, and Com-Streams with channel widths $C_h$, $C_h-C_c$, and $C_c$, respectively, where $C_c=\lfloor eD\rfloor$ and $e=1.0$ by default. This heterogeneous allocation allows only a controlled subset of channels to undergo stronger computation while retaining a sufficiently wide information-preserving pathway. In shallow stages, HSG-WAVE performs wavelet-based frequency decoupling to strengthen low-frequency structural modeling while preserving fine-grained details. In deeper stages, HSG-AKAT equips the computation stream with asymmetric self-attention and Group KAN, enabling more expressive global semantic interaction and selective nonlinear abstraction.

\subsubsection{HSG-WAVE}
CNNs exhibit strong locality and translation-equivariant biases, which make shallow layers effective at capturing edges, textures, and other high-frequency patterns. However, these local responses often dominate early representations, while low-frequency structural cues related to object shape, regional consistency, and global layout remain insufficiently exploited. We argue that this weakness stems not only from limited receptive fields, but also from the spectral entanglement between dominant high-frequency details and weaker low-frequency structures. To mitigate this issue, we introduce HSG-WAVE, which explicitly decouples shallow features into low-frequency structural components and high-frequency detail components. By recursively modeling only the low-frequency branch, HSG-WAVE enhances the network’s ability to perceive and exploit global structural cues while preserving fine-grained details.

Given an input feature \(X_{p}^{c}\), we first apply the Haar Wavelet Transform (HWT) to decompose it into one low-frequency sub-band and three high-frequency sub-bands:
\begin{equation}
\label{eq:hwt_decomp}
\mathcal{W}(X_{p}^{c(l)})=
\{X_{LL}^{(l)},X_{LH}^{(l)},X_{HL}^{(l)},X_{HH}^{(l)}\},
\end{equation}
where \( \mathcal{W}(\cdot) \) denotes the HWT. 
\(X_{LL}^{(l)}\) encodes coarse structural information, while \(X_{LH}^{(l)}\), \(X_{HL}^{(l)}\), and \(X_{HH}^{(l)}\) capture horizontal, vertical, and diagonal details, respectively. For simplicity, we denote the four sub-bands at level \(l\) as 
\(\mathcal{S}^{(l)}=\{X_{LL}^{(l)},X_{LH}^{(l)},X_{HL}^{(l)},X_{HH}^{(l)}\}\). To enlarge the effective receptive field (ERF) with low computational cost, HSG-WAVE recursively applies HWT only to the low-frequency branch. Then, at each decomposition level, the four sub-bands are processed by a grouped convolution and then transformed back to the spatial domain to generate a coarser structural prior:
\begin{equation}
\label{eq:recursive_prior}
R^{(l+1)}
=
\mathcal{W}^{-1}
\left(
\mathcal{G}_{5\times5}
\left(
\mathcal{S}^{(l+1)}
\right)
\right),
\end{equation}
where \( \mathcal{G}_{5\times5}(\cdot) \) denotes the grouped convolution and \( \mathcal{W}^{-1}(\cdot) \) denotes the Inverse HWT. Since the recursively decomposed \(LL\) branch has reduced spatial resolution, structural context modeling introduces only marginal computational overhead. The generated structural prior is used to refine the low-frequency component at the current level, and the enhanced feature is reconstructed as:
\begin{equation}
\label{eq:hsg_wave_reconstruct}
\begin{aligned}
X_{p}^{c(l)}
=
\mathcal{W}^{-1}
\big(
& X_{LL}^{(l)}+R^{(l+1)}, X_{LH}^{(l)}, X_{HL}^{(l)}, X_{HH}^{(l)}
\big).
\end{aligned}
\end{equation}
Through this recursive analysis-and-synthesis process, HSG-WAVE injects global structural semantics into shallow features while retaining fine local details. Compared with conventional convolutions that tend to overemphasize local high-frequency responses, the proposed module improves shallow representations by better coordinating holistic structural perception and fine-grained detail preservation, while suppressing the dominance of unstable high-frequency noise.

\subsubsection{HSG-AKAT}
As the network depth increases, the representation focus gradually shifts from structure-detail disentanglement to semantic abstraction and long-range dependency modeling. In the deeper stages of the network, the Transformer's global modeling capacity becomes increasingly valuable. However, standard Transformer blocks face two challenges: i) self-attention is computationally expensive and often disrupts the intrinsic 2D spatial structure when flattening features into 1D token sequences; ii) conventional MLPs are static and data-agnostic, which limits their ability to capture complex nonlinear contextual dependencies \cite{liu2024kan}. Therefore, in the deeper stages, we instantiate the Com-Stream with the Asymmetric Kolmogorov-Arnold Transformer (AKAT), an expressive module designed to enhance global semantic interaction and nonlinear contextual representation.

\paragraph{Asymmetric Self-Attention}
Given an input feature map $X_{p}^{c} \in \mathbb{R}^{C \times H \times W}$, we first generate query, key, and value projections by lightweight $1 \times 1$ convolutions. Let $N=HW$ denote the number of spatial tokens and $h$ the number of attention heads. After flattening and head-wise partitioning, we obtain $Q,K\in\mathbb{R}^{h\times N\times d_k}$ and $V\in\mathbb{R}^{h\times N\times d_v}$, where $d_k<d_v$. In our implementation, $h=8$ and $d_k=0.5d_v$. This asymmetric projection performs the expensive query--key interaction in a lower-dimensional subspace while retaining a richer value representation for semantic aggregation.

To maintain local spatial awareness, we further derive a value-dependent local positional encoding from the value tensor:
\begin{equation}
Pos=
\mathrm{Flatten}\!\left(
\mathrm{DWConv}_{3\times3}
\left(\mathrm{Reshape}(V)\right)
\right),
\end{equation}
where $\mathrm{DWConv}$ denotes depthwise convolution. After head-wise partitioning, $Pos\in\mathbb{R}^{h\times N\times d_v}$ has the same dimension as the aggregated value feature. Specifically, $QK^\top\in\mathbb{R}^{h\times N\times N}$, and its multiplication with $V$ produces a feature in $\mathbb{R}^{h\times N\times d_v}$. The attention output is therefore computed as
\begin{equation}
\mathrm{Attn}(Q,K,V)
=
\mathrm{Softmax}\!\left(
\frac{QK^\top}{\sqrt{d_k}}
\right)V+Pos.
\end{equation}
This design enables efficient global dependency modeling while preserving local spatial inductive information, making it suitable for deep semantic refinement within the Computation Stream.

\paragraph{Group KAN}
Instead of using a conventional MLP with fixed pointwise activations after attention, we employ a grouped Kolmogorov--Arnold transformation to enhance nonlinear feature modeling within the deep Com-Stream. Specifically, the channels are uniformly partitioned into \(G\) groups, and each group is equipped with an independently learnable rational function
\begin{equation}
\mathcal{R}_{g}(x)
=
\frac{P_{g}(x)}{Q_{g}(x)},
\qquad g=1,\ldots,G,
\end{equation}
where \(P_g(\cdot)\) and \(Q_g(\cdot)\) denote learnable numerator and denominator polynomials, respectively. In our implementation, we use \(G=8\) groups with rational-function orders \((5,4)\), providing group-specific nonlinear mappings while keeping the parameter and computational overhead moderate.

The Group KAN transformation follows a two-layer channel projection with an expansion ratio of four:
\begin{equation}
F_{\mathrm{KAN}}
=
W_2\,
\mathcal{R}_{\mathrm{GELU}}
\left(
W_1\,
\mathcal{R}_{\mathrm{Id}}(F)
\right),
\end{equation}
where \(W_1:C\rightarrow4C\) and \(W_2:4C\rightarrow C\) denote linear projections. \(\mathcal{R}_{\mathrm{Id}}\) and \(\mathcal{R}_{\mathrm{GELU}}\) are grouped learnable rational mappings initialized from identity and GELU functions, respectively, and are subsequently optimized together with the network. Compared with a conventional MLP using a fixed activation function, this formulation provides adaptive group-wise nonlinear transformations and improves the expressive capacity of high-level semantic features without introducing dense channel-wise functional parameterization.

Combined with asymmetric self-attention, Group KAN enables HSG-AKAT to jointly capture long-range semantic dependencies and adaptive nonlinear feature interactions within the deep Com-Stream. Overall, WeKat establishes a stage-aware representation hierarchy through the HSG design. By separating feature propagation into retention, gating, and computation streams, HSG preserves stable information pathways while selectively enhancing informative channels. HSG-WAVE emphasizes structure--detail coordination in shallow stages, whereas HSG-AKAT strengthens global semantic interaction and nonlinear representation in deeper stages. This progressive design enables WeKat to balance fine-grained detail preservation, structural consistency, and high-level semantic modeling for high-resolution RSOD.
\subsection{Multi-Domain Feature Coordination}
\begin{figure*}[t]
\centering
\includegraphics[width=.9\textwidth]{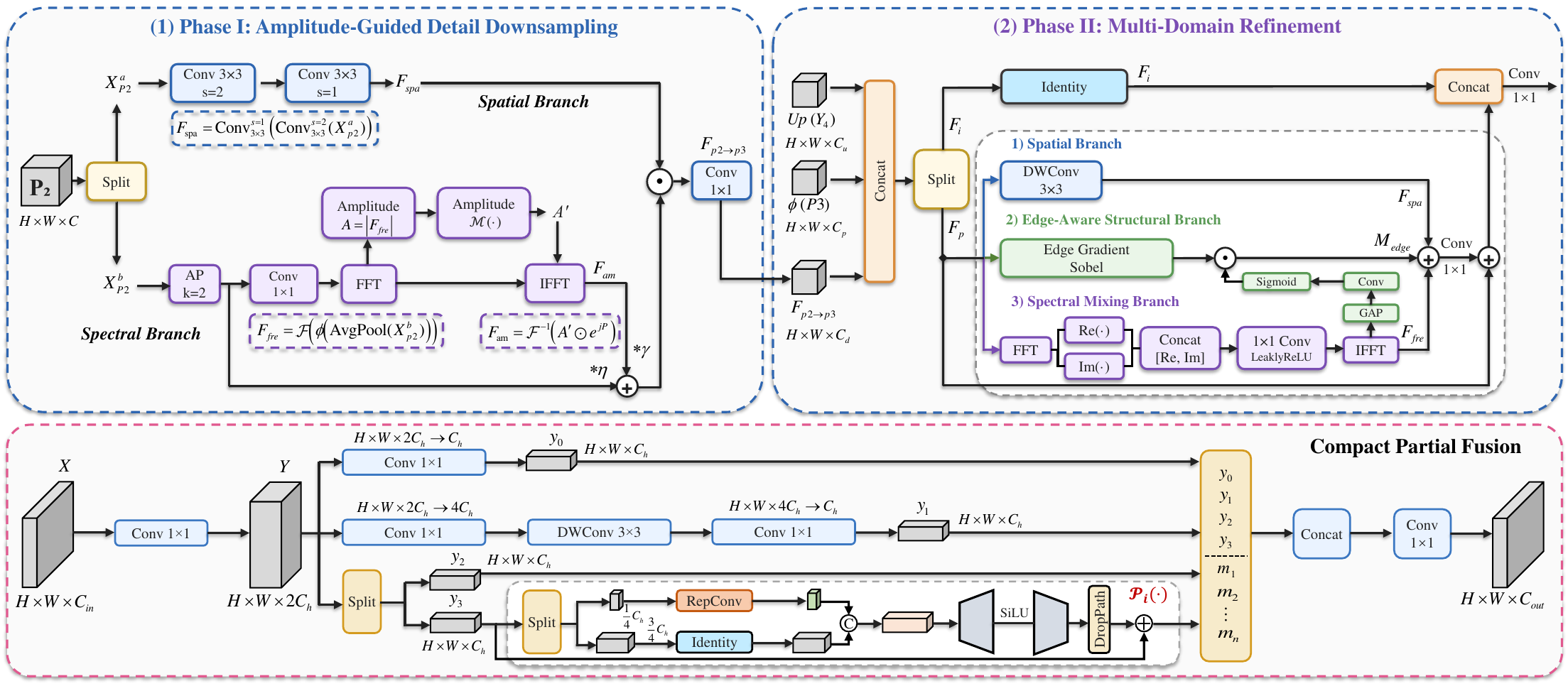} 
\caption{Overview of the proposed MDFC and CPF modules. The upper part shows the MDFC module, which consists of two successive phases: (1) amplitude-guided detail downsampling and (2) multi-domain refinement. The lower part illustrates the Compact Partial Fusion (CPF) module.}
\label{fig3_mdfc}
\end{figure*}

In remote sensing imagery, small targets usually occupy only a few pixels, making shallow high-resolution features crucial for capturing fine-grained local structures \cite{liu2025dmsa}. Furthermore, after cross-scale concatenation, the aggregated representation still suffers from substantial heterogeneity, since shallow features mainly provide local details while deeper features contribute stronger semantic context. To this end, we introduce the Multi-Domain Feature Coordination (MDFC) module as a cross-scale coordination block. It couples shallow detail injection with partial-channel multi-domain refinement to calibrate aggregated features from local spatial, global spectral, and edge-structural perspectives, thereby enhancing tiny-object cues while suppressing background interference. As shown in Fig.~\ref{fig3_mdfc}, MDFC is composed of two successive phases: amplitude-guided detail downsampling and multi-domain refinement.

In the first phase, the shallow feature \(P_2\) is transformed into a detail-enhanced prior aligned with the \(P_3\) resolution through an amplitude-modulated fine-grained downsampling unit. Specifically, the input feature is evenly split into a spatial branch \(X_{p2}^{a}\) and a spectral branch \(X_{p2}^{b}\). The spatial branch is processed by two successive convolutions,
\begin{equation}
F_{\mathrm{spa}}=\mathrm{Conv}_{3\times3}^{s=1}\!\left(\mathrm{Conv}_{3\times3}^{s=2}(X_{p2}^{a})\right),
\end{equation}
which reduce the spatial resolution while preserving local structural continuity. In parallel, the second branch is fed into an amplitude-modulated spectral pathway. After average pooling and channel alignment, we define
\begin{equation}
Z=\phi\!\left(\mathrm{AvgPool}(X_{p2}^{b})\right), \qquad
F_{\mathrm{fre}}=\mathcal{F}(Z),
\end{equation}
where $\phi(\cdot)$ denotes a $1\times1$ convolution and $\mathcal{F}(\cdot)$ denotes the Fast Fourier Transform (FFT). Then, $F_{\mathrm{fre}}$ is explicitly decomposed into amplitude and phase:
\begin{equation}
A=\left|F_{\mathrm{fre}}\right|,\qquad
P=\angle F_{\mathrm{fre}}.
\end{equation}
We apply learnable convolutional modulation only to the amplitude,
\begin{equation}
A'=\mathcal{M}(A),
\end{equation}
where \(\mathcal{M}(\cdot)\) denotes the 1$\times$1 convolution operator, while preserving the original phase to maintain structural consistency. The amplitude-modulated response is reconstructed with the original phase and combined with the aligned spectral feature:
\begin{equation}
F_{\mathrm{am}}=\mathcal{F}^{-1}\!\left(A'\odot e^{jP}\right), \qquad
F_{\mathrm{spe}}=\gamma F_{\mathrm{am}}+\eta Z,
\end{equation}
where $\mathcal{F}^{-1}(\cdot)$ denotes the inverse Fast Fourier Transform (IFFT), and $\gamma$ and $\eta$ are learnable scaling coefficients for the modulated and residual spectral responses, respectively. Finally, the spatial and spectral responses are coupled through element-wise interaction:
\begin{equation}
F_{p2\rightarrow p3}
=
\phi\!\left(
F_{\mathrm{spa}}\odot F_{\mathrm{spe}}
\right).
\end{equation}
In this way, shallow details are not injected in raw form, but selectively reweighted by frequency-aware amplitude cues before being aligned to the \(P_3\) scale.

In the second phase, the generated shallow detail prior is aggregated with the adjacent multi-scale features:
\begin{equation}
F_{\mathrm{cat}}=\mathrm{Concat}\left[\mathrm{Up}(Y_4),\,\phi(P_3),\,F_{p2\rightarrow p3}\right],
\end{equation}
where \(\mathrm{Up}(\cdot)\) denotes the upsampling operator. The concatenated representation is then refined by a multi-domain refinement unit. Within the refinement branch, spatial, spectral, and edge-aware structural responses are jointly modeled and aggregated. First, the feature \(F_{cat}\) is split into a process branch \(F_p\) and an identity branch \(F_i\) with a 1:3 channel ratio, where \(F_p\) undergoes heavy transformation while \(F_i\) preserves low-perturbation information flow. A spatial branch captures local patterns by depthwise convolution:
\begin{equation}
F_{\mathrm{spa}}=\mathrm{DWConv}_{3\times3}(F_p).
\end{equation}
A spectral mixing branch performs global frequency-domain interaction. Specifically, we transform \(F_p\) into the frequency domain and concatenate its real and imaginary parts:
\begin{equation}
\widehat{F}_{p}=\mathcal{F}(F_p),\qquad
\widehat{F}_{\mathrm{ri}}=[\Re(\widehat{F}_{p}),\,\Im(\widehat{F}_{p})].
\end{equation}
The concatenated spectral response is then modulated by a learnable \(1\times1\) convolution and mapped back to the spatial domain:
\begin{equation}
F_{\mathrm{fre}}=\left|\mathcal{F}^{-1}\!\left(\Gamma(\widehat{F}_{\mathrm{ri}})\right)\right|,
\end{equation}
where \(\Gamma(\cdot)\) denotes a \(1\times1\) convolution followed by LeakyReLU activation. The resulting real and imaginary components are recombined into a complex spectrum before the inverse Fourier transform. To further inject explicit structural priors, we derive an edge-aware response by using the frequency-enhanced feature to gate Sobel gradients:
\begin{equation}
M_{\mathrm{edge}}=\sigma\!\left(\delta(\mathrm{GAP}(F_{\mathrm{fre}}))\right)\odot \mathrm{Sobel}(F_p),
\end{equation}
where \(\mathrm{GAP}(\cdot)\) denotes global average pooling, \(\delta(\cdot)\) is a learnable convolution transform, and \(\sigma(\cdot)\) denotes the sigmoid function. The three responses are then jointly aggregated:
\begin{equation}
F_{\mathrm{join}}=F_p+\phi\!\left(F_{\mathrm{spa}}+F_{\mathrm{fre}}+M_{\mathrm{edge}}\right).
\end{equation}
Finally, the refined process branch is concatenated with the identity branch and fused by a \(1\times1\) convolution:
\begin{equation}
F_{\mathrm{MDFC}}=\phi\!\left(\mathrm{Concat}[F_{\mathrm{join}},\,F_i]\right).
\end{equation}

Overall, the MDFC is a cross-scale multi-domain coordination module that first converts shallow low-level features into an amplitude-guided detail prior and then refines the aggregated representation through partial-channel spatial, spectral, and structural modeling. This design is particularly beneficial for small object detection, as it strengthens the fine-grained representation with complementary local detail, global frequency context, and explicit edge cues, while preserving computational efficiency and avoiding excessive disturbance to all channels.

\subsection{Compact Partial Fusion}
Effective feature fusion and interaction are essential for improving representation quality, especially in complex scenarios that require both sufficient local detail modeling and stable feature propagation \cite{xiao2025fbrt,yang2025mstdfgrn,yang2025mtegcrn}. To enhance feature diversity while avoiding excessive feature perturbation, we propose Compact Partial Fusion (CPF), a compact multi-branch aggregation module with progressive partial refinement.

As illustrated in Fig.~\ref{fig3_mdfc}, given an input feature map
\(X \in \mathbb{R}^{H \times W \times C_{\mathrm{in}}}\),
CPF first projects it into an intermediate representation
\(Y \in \mathbb{R}^{H \times W \times 2C_h}\)
through a \(1\times1\) convolution, where
\(C_h=0.5C_{\mathrm{out}}\).
Based on \(Y\), four complementary feature branches are constructed.
The first branch applies a lightweight \(1\times1\) convolution to produce \(y_0\), providing direct channel interaction. The second branch adopts a pointwise--depthwise--pointwise transformation with channel dimensions
\(2C_h \rightarrow 4C_h \rightarrow C_h\)
to generate \(y_1\), which enhances local contextual responses with a larger spatial mixing capacity. Meanwhile, \(Y\) is evenly split into two parts, denoted as \(y_2\) and \(y_3\), to preserve the projected features and maintain stable information flow:
\begin{equation}
y_0 = \phi_0(Y), \quad
y_1 = \phi_1(Y), \quad
(y_2,y_3)=\mathrm{Split}(Y),
\end{equation}
where
\(y_0,y_1,y_2,y_3 \in \mathbb{R}^{H\times W\times C_h}\),
\(\phi_0(\cdot)\) is a \(1\times1\) convolution, and
\(\phi_1(\cdot)\) denotes the pointwise--depthwise--pointwise branch. To further refine informative responses, CPF progressively processes \(y_3\) through a sequence of partial re-parameterized refinement blocks:
\begin{equation}
m_i = \mathcal{P}_i(m_{i-1}), \quad i=1,\dots,n,
\end{equation}
where \(m_0=y_3\), \(\mathcal{P}_i(\cdot)\) denotes the $i$-th refinement block, and \(n=3\) in our implementation. Within each refinement block, the channels are divided at a ratio of \(1:3\): only \(C_h/4\) channels are transformed by a re-parameterized \(3\times3\) convolution, while the remaining \(3C_h/4\) channels are preserved through an identity path. The two parts are then concatenated and passed through a lightweight channel transformation with an expansion ratio of 2, i.e.,
\(C_h \rightarrow 2C_h \rightarrow C_h\),
together with a residual connection. This partial refinement strategy improves spatial representation while avoiding unnecessary disturbance to all feature channels. Finally, CPF aggregates the directly transformed branches, preserved features, and progressively refined features through concatenation followed by a \(1\times1\) fusion convolution:
\begin{equation}
\hat{X}
=
\phi\left(
\mathrm{Concat}
\left(
y_0,y_1,y_2,y_3,m_1,\dots,m_n
\right)
\right).
\end{equation}

Compared with single-path aggregation, CPF introduces a more compact yet diverse feature interaction pattern. The direct transformation branches enhance channel and local contextual modeling, the preserved branches maintain stable feature propagation, and the partial refinement path progressively strengthens informative spatial responses. As a result, CPF improves feature aggregation capability while retaining a stable information pathway, which is beneficial for robust object detection in complex scenes.
\section{Experiments}
In this section, we first introduce the RSOD datasets used in our experiments, including VisDrone \cite{cao2021visdrone}, HazyDet \cite{feng2026hazydet}, and SIMD \cite{haroon2020multisized}. We then describe the experimental settings and present comparisons with state-of-the-art methods. Subsequently, ablation studies are conducted to evaluate the effectiveness of each component, followed by qualitative visualizations for further analysis.

\subsection{Datasets}
\subsubsection{VisDrone}
VisDrone is a large-scale benchmark specifically designed for object detection in aerial imagery. It consists of high-resolution UAV-captured images from diverse urban and suburban scenes across 14 cities in China, covering a wide range of real-world scenarios. The dataset contains 6,471 training, 548 validation, and 3,190 test images, with over 2.5 million bounding boxes annotated across ten common object categories. Due to the prevalence of small objects, severe occlusions, crowded scenes, and complex backgrounds, VisDrone presents significant challenges for detection algorithms, making it a valuable benchmark for evaluating the robustness and generalization of aerial object detection models.

\subsubsection{HazyDet} 
HazyDet is the first large-scale benchmark dedicated to object detection in hazy aerial imagery. It contains 11,000 high-resolution images collected from both real-world scenes and physics-based simulations, with 383,000 bounding boxes annotated across three vehicle categories: car, truck, and bus. The dataset is split into 8,000 training, 1,000 validation, and 2,000 test images. By incorporating naturally captured hazy images and synthetic data with varying visibility levels, HazyDet fills a critical gap left by clear-weather benchmarks and provides a valuable resource for evaluating the robustness of detection models under adverse weather conditions.

\subsubsection{SIMD}
SIMD is a medium-scale dataset specifically designed for multi-scale and multi-class vehicle detection in satellite imagery. It consists of 5,000 RGB images with a resolution of 1024$\times$768, collected from 79 locations across Europe and the United States via Google Earth, and adopts a 4:1 training-to-testing split. In total, it provides 45,096 annotated objects spanning 15 categories, mainly vehicles such as cars, trucks, buses, and long vehicles, as well as multiple aircraft types and boats. The diversity in object appearances, scales, densities, and backgrounds makes SIMD a valuable benchmark for advancing aerial object detection, surveillance, and automatic scene analysis under realistic conditions.

\subsection{Implementation Details}
All experiments are conducted using PyTorch 2.3 with Python 3.10 on an NVIDIA GeForce RTX 4090D GPU (24 GB). To ensure fair and reproducible comparisons, our method is built upon the RT-DETR \cite{zhao2024detrs} baseline and trained from scratch without using any pretrained weights or multi-scale training strategies. Mosaic augmentation is employed during training to improve data diversity and robustness. Unless otherwise specified, the input resolution is fixed at 640$\times$640 during both training and testing. The network is trained for 200 epochs using the AdamW optimizer, with a momentum of 0.9, a weight decay of 0.0005, a batch size of 4, and an initial learning rate of 0.0001. We evaluate performance primarily using the COCO-style Average Precision (AP), along with additional metrics at different IoU thresholds and object scales, including AP$_{50}$, AP$_{75}$, and AP$_{S}$. Throughout this paper, FMC-DETR denotes the overall framework. FMC-DETR-B (Base) is the default general configuration using $[D3,D4,D5]$, while FMC-DETR-T (TinyAware) is an additional small-object-oriented variant derived from FMC-DETR-B by removing $S5$ and adopting $[D2,D4]$.

\begin{table*}[!t]
    \caption{Quantitative comparison with state-of-the-art detectors on the VisDrone validation set. \textbf{FMC-DETR-B (FMC-DETR-Base)} denotes the default general configuration, and \textbf{FMC-DETR-T (FMC-DETR-TinyAware)} denotes the additional small-object-oriented variant. Results marked with $^{\dagger}$ are reproduced by us under the same experimental settings. The best and second-best results are highlighted in \textcolor{red}{red} and \textcolor{blue}{blue}, respectively.}
    \label{table1_visdrone}
    \centering
    \setlength{\tabcolsep}{4mm}
    \begin{tabular}{c|ccccccc}
    \toprule
    \textbf{Method} & \textbf{Publication} & \textbf{Input size} & \textbf{AP} & \textbf{AP$_{50}$} & \textbf{AP$_S$} & \textbf{Param} & \textbf{FLOPs} \\
    \hline
    \rowcolor{gray!15}
    \multicolumn{8}{l}{\textit{\textbf{CNN-Based}}} \\
    \hline
    YOLOv8-L \cite{yolov8_ultralytics} & Ultralytics & 640$\times$640 & 28.4 & 45.9 & -- & 43.7M & 165.2G \\
    YOLOv8-X \cite{yolov8_ultralytics} & Ultralytics & 640$\times$640 & 28.9 & 46.8 & -- & 68.2M & 257.8G \\
    YOLOv9-M \cite{wang2024yolov9} & ECCV2024 & 640$\times$640 & 25.1 & 41.9 & -- & 20.0M & 76.3G \\
    YOLOv10-L \cite{wang2024yolov10} & NeurIPS2024 & 640$\times$640 & 27.6 & 44.6 & -- & 24.4M & 120.3G \\
    YOLOv10-X \cite{wang2024yolov10} & NeurIPS2024 & 640$\times$640 & 28.7 & 46.1 & -- & 29.5M & 160.4G \\
    YOLOv11-M $^{\dagger}$ \cite{yolo11_ultralytics} & Ultralytics & 640$\times$640 & 25.0 & 42.0 & 15.7 & 20.1M & 68.0G \\
    YOLOv11-L $^{\dagger}$ \cite{yolo11_ultralytics} & Ultralytics & 640$\times$640 & 25.5 & 42.2 & 15.9 & 25.3M & 86.9G \\
    YOLOv11-X $^{\dagger}$ \cite{yolo11_ultralytics} & Ultralytics & 640$\times$640 & 26.6 & 43.8 & 17.0 & 56.9M & 194.9G \\
    YOLOv12-L $^{\dagger}$ \cite{tian2025yolov12} & NeurIPS2025 & 640$\times$640 & 25.1 & 42.0 & 15.2 & 26.4M & 88.9G \\
    YOLOv13-L $^{\dagger}$ \cite{lei2025yolov13} & arXiv2025 & 640$\times$640 & 24.2 & 40.5 & 14.5 & 27.6M & 88.4G \\
    YOLO26-L $^{\dagger}$ \cite{yolo11_ultralytics} & arXiv2025 & 640$\times$640 & 25.7 & 43.0 & 16.7 & 24.8M & 86.4G \\
    YOLO-Master-L $^{\dagger}$ \cite{lin2025yolo} & CVPR2026 & 640$\times$640 & 25.9 & 42.8 & 15.8 & 58.4M & 138.1G \\
    DTSSNet \cite{chen2024dtssnet} & TGRS2024 & 640$\times$640 & 24.2 & 39.9 & 17.5 & 10.1M & 49.6G \\
    FBRT-YOLO-M \cite{xiao2025fbrt} & AAAI2025 & 640$\times$640 & 28.4 & 45.9 & -- & 7.2M & 58.7G \\
    FBRT-YOLO-L \cite{xiao2025fbrt} & AAAI2025 & 640$\times$640 & 29.7 & 47.7 & -- & 14.6M & 119.2G \\
    FBRT-YOLO-X \cite{xiao2025fbrt} & AAAI2025 & 640$\times$640 & \textbf{\textcolor{blue}{30.1}} & \textbf{\textcolor{blue}{48.4}} & -- & 22.8M & 185.8G \\
    \hline
    \rowcolor{gray!15}
    \multicolumn{8}{l}{\textit{\textbf{Transformer-Based}}} \\
    \hline
    Deform DETR \cite{zhu2020deformable} & ICLR2021 & 1300$\times$800 & 27.1 & 42.2 & -- & 40.0M & 173.0G \\
    Sparse DETR \cite{roh2021sparse} & ICLR2022 & 1300$\times$800 & 27.3 & 42.5 & -- & 40.9M & 121.0G \\
    RT-DETR-R18 $^{\dagger}$ \cite{zhao2024detrs} & CVPR2024 & 640$\times$640 & 26.7 & 44.6 & 18.5 & 20.0M & 60.0G \\
    RT-DETR-R50 $^{\dagger}$ \cite{zhao2024detrs} & CVPR2024 & 640$\times$640 & 28.4 & 47.0 & 20.7 & 42.0M & 136.0G \\
    Mamba-YOLO-T $^{\dagger}$ \cite{wang2025mamba} & AAAI2025 & 640$\times$640 & 21.0 & 36.8 & 11.6 & \textbf{\textcolor{blue}{6.0M}} & \textbf{\textcolor{blue}{13.6G}} \\
    Mamba-YOLO-B $^{\dagger}$ \cite{wang2025mamba} & AAAI2025 & 640$\times$640 & 23.9 & 40.8 & 14.8 & 21.8M & 49.6G \\
    DEIM-D-FINE-N $^{\dagger}$ \cite{huang2025deim} & CVPR2025 & 640$\times$640 & 17.8 & 31.5 & 11.3 & \textbf{\textcolor{red}{3.7M}} & \textbf{\textcolor{red}{7.1G}} \\
    DEIM-D-FINE-S $^{\dagger}$ \cite{huang2025deim} & CVPR2025 & 640$\times$640 & 24.3 & 40.6 & 16.9 & 10.1M & 24.9G \\
    \hline
    \rowcolor{cyan!12}
    FMC-DETR-B (Ours) & - & 640$\times$640 & 29.4 & \textbf{\textcolor{blue}{48.4}} & \textbf{\textcolor{blue}{21.4}} & 17.4M & 62.9G \\
    \rowcolor{cyan!12}
    FMC-DETR-T (Ours) & - & 640$\times$640 & \textbf{\textcolor{red}{33.7}} & \textbf{\textcolor{red}{53.6}} & \textbf{\textcolor{red}{25.8}} & 13.8M & 146.8G \\
    \bottomrule
    \end{tabular}
\end{table*}
\subsection{Comparison With State-of-the-Art Methods}
\subsubsection{Results on VisDrone Dataset}
The results in Table~\ref{table1_visdrone} demonstrate the superior performance of the proposed method on the VisDrone validation set. FMC-DETR-B achieves an AP of 29.4\% and an AP$_{50}$ of 48.4\%, outperforming the baseline RT-DETR by 2.7\% AP and 3.8\% AP$_{50}$, respectively. Furthermore, FMC-DETR-T adopts a compact two-layer detection configuration, \([D2,D4]\), to exploit the higher-resolution \(D2\) feature. It achieves an AP of 33.7\% and an AP$_{50}$ of 53.6\%, demonstrating strong performance among the compared detectors. Compared with recent YOLO-based detectors, including YOLOv12-L, YOLOv13-L, YOLO26-L, and YOLO-Master-L, FMC-DETR-T consistently achieves higher AP$_{50}$ scores, with gains of 11.6, 13.1, 10.6, and 10.8 points, respectively. Notably, FBRT-YOLO is a recent representative detector specifically developed for aerial small object detection. Compared with its strongest variant, FBRT-YOLO-X, FMC-DETR-T improves AP and AP$_{50}$ by 3.6 and 5.2 points, respectively, while using fewer parameters and lower computational cost. Compared with the Transformer-based baseline RT-DETR-R18, FMC-DETR-T improves AP from 26.7\% to 33.7\% and AP$_{50}$ from 44.6\% to 53.6\%, corresponding to gains of 7.0 and 9.0 percentage points, respectively. In addition, the general variant FMC-DETR-B achieves 29.4\% AP and 48.4\% AP$_{50}$ with 17.4M parameters and 62.9G FLOPs, showing improved detection accuracy while maintaining a moderate number of parameters.

\begin{table}[t!]
    \caption{Comparison of the performance of different SOTA detectors on the HazyDet dataset.}
    \label{table2_hazydet}
    \centering
    \small
    \setlength{\tabcolsep}{1.5mm}
    \begin{tabular}{l|cccc|c}
    \hline
    \textbf{Model}   & \textbf{AP}  & \textbf{AP$^{car}$}   & \textbf{AP$^{truck}$}   & \textbf{AP$^{bus}$}  & \textbf{Param} \\
    \hline
    RT-DETR-R18      & 53.3   & 62.4 & 36.3 & 61.2  & 20.0M \\
    IAYOLO           & 38.3   & 44.1 & 22.2 & 48.6  & 61.8M \\
    MS-DAYOLO        & 48.3   & 59.4 & 28.5 & 57.0  & 40.0M \\
    TOOD             & 51.4   & 58.4 & 33.6 & 62.2  & 32.0M \\
    Cascade RCNN     & 51.6   & 59.0 & 34.2 & 61.7  & 69.1M \\
    Deform DETR     & 51.5   & 58.4 & 33.9 & 62.3  & 40.0M \\
    DeCoDet          & 52.0   & 60.5 & 34.0 & 61.9  & 34.6M \\
    YOLOv12-L        & 52.6   & 57.5 & 37.4 & 62.6  & 26.4M \\
    YOLOv13-L        & 52.7   & 57.5 & 37.6 & 63.1  & 27.6M \\
    DEIM-D-FINE-S    & 53.5   & 62.5 & 34.0 & 64.0  & 10.1M \\
    FBRT-YOLO-X      & 55.0   & 59.4 & 41.0 & 64.8 & 22.8M \\ 
    \rowcolor{cyan!12}
    FMC-DETR-B       & 55.0   & 63.7  & 37.6 & 63.5  & 17.4M \\  
    \hline
    \end{tabular}
\end{table}

\begin{table}[t!]
    \caption{Comparison of the performance of different SOTA detectors on the SIMD dataset.}
    \label{table3_SIMD}
    \centering
    \small
    \setlength{\tabcolsep}{3.0mm}
    \begin{tabular}{l| cc| cc}
    \hline
    \textbf{Model}   & \textbf{AP}  & \textbf{AP$_{50}$}    & \textbf{Param} & \textbf{FLOPs}  \\
    \hline
    RT-DETR-R18      & 63.7  & 78.6      & 20.0M  & 60.0G \\
    YOLOv8-L         & 63.1  & 78.1      & 43.7M  & 165.2G \\
    YOLOv9-M         & 62.2  & 76.6      & 20.0M  & 76.3G \\
    DEIM-D-FINE-S    & 66.2  & 82.0      & 10.1M  & 24.9G \\
    FBRT-YOLO-X      & 66.2  & 83.0      & 22.8M  & 185.8G \\
    Deform DETR      & 59.7  & 75.6      & 40.0M  & 173.0G \\
    EMSD-DETR        & 64.3  & 79.4      & 18.4M  & 68.3G \\
    HPS-DETR         & 63.5  & 79.8      & 15.5M  & 68.3G \\
    \rowcolor{cyan!12}
    FMC-DETR-B       & 66.6  & 80.9    & 17.4M  & 62.9G  \\
    \hline
    \end{tabular}
\end{table}

\subsubsection{Results on HazyDet Dataset}
We compare FMC-DETR-B against several representative object detectors on the challenging HazyDet dataset. As shown in Table~\ref{table2_hazydet}, FMC-DETR-B achieves the best overall AP of 55.0\% among the compared methods, outperforming strong competing methods such as DeCoDet~\cite{feng2026hazydet} (52.0\%), YOLOv12-L (52.6\%), and DEIM-D-FINE-S (53.5\%). These results underscore the superior detection capability of our method in complex aerial-view scenarios. In terms of category-wise performance, FMC-DETR-B achieves a leading 63.7\% AP on the Car class, which dominates the dataset and primarily consists of small, densely packed targets—highlighting the model’s effectiveness in capturing fine-grained details under low-visibility conditions.

\subsubsection{Results on SIMD Dataset} 
We evaluate FMC-DETR-B on the SIMD dataset and compare its performance with several recent object detection models. As shown in Table~\ref{table3_SIMD}, FMC-DETR-B achieves an AP of 66.6\%, outperforming models such as RT-DETR-R18 (63.7\%), YOLOv8-L (63.1\%), and YOLOv9-M (62.2\%), while maintaining a relatively low parameter count of 17.4M and FLOPs of 62.9G. Compared with improved DETR-based models, FMC-DETR-B achieves a higher AP$_{50}$ than Deform DETR \cite{zhu2020deformable}, EMSD-DETR \cite{zhang2025emsd}, and HPS-DETR \cite{wang2025hps} by 5.3\%, 1.5\%, and 1.1\%, respectively.
\subsection{Ablation Studies}
To validate the effectiveness of our proposed modules, we conducted comprehensive ablation studies across multiple datasets, including VisDrone \cite{cao2021visdrone}, HazyDet \cite{feng2026hazydet}, and SIMD \cite{haroon2020multisized}, as shown in the following tables.

\subsubsection{Effectiveness of the Proposed Modules Across Datasets}
To disentangle the contributions of the proposed modules from those of the detection-layer design, we first conduct a module-level ablation under a fixed detection-layer configuration. Tables~\ref{table4_visdrone_ablation} and~\ref{table4_HazyDet_and_SIMD_Ablation} present the ablation studies of WeKat, MDFC, and CPF. On VisDrone, all three modules consistently improve the RT-DETR-R18 baseline. Notably, WeKat increases AP from 26.7\% to 27.8\% while reducing the parameters and FLOPs from 20.0M/60.0G to 16.1M/53.3G, indicating that the proposed backbone improves feature representation with lower model complexity. Among the individual modules, MDFC achieves the largest gain, reaching 28.2\% AP and 46.7\% AP$_{50}$, which verifies the importance of multi-domain feature coordination for aerial small objects. CPF also brings consistent improvements, showing that compact partial aggregation is beneficial for enhancing local feature interaction and stable information fusion. By integrating all three modules, the full model achieves the best performance, with 29.4\% AP, 48.4\% AP$_{50}$, and 21.4\% AP$_S$, corresponding to improvements of 2.7, 3.8, and 2.9 points over the baseline, respectively. The cross-dataset results on HazyDet and SIMD further validate the generalization ability of the proposed components. 
On HazyDet, the complete model improves AP/AP$_{50}$/AP$_{75}$ from 53.3/73.6/62.2 to 55.0/74.5/63.6~($\uparrow$1.7/$\uparrow$0.9/$\uparrow$1.4). On SIMD, the gains are more pronounced, with AP/AP$_{50}$/AP$_{75}$ increasing from 63.7/78.6/72.4 to 66.6/80.9/77.8~($\uparrow$2.9/$\uparrow$2.3/$\uparrow$5.4). These consistent improvements across different datasets demonstrate that WeKat, MDFC, and CPF provide complementary benefits and jointly enhance feature representation and detection robustness in aerial object detection scenarios.

\begin{table}[!t]
    \centering
    \small
    \caption{Ablation studies on the VisDrone dataset.}
    \label{table4_visdrone_ablation}
    \setlength{\tabcolsep}{1.2mm}
    \begin{tabular}{ccc|ccc|cc}
    \toprule
    \textbf{WeKat} & \textbf{MDFC} & \textbf{CPF} & \textbf{AP} & \textbf{AP$_{50}$} & \textbf{AP$_{S}$} & \textbf{Param} & \textbf{FLOPs} \\
    \midrule

    \ding{55} & \ding{55} & \ding{55}   & 26.7 & 44.6 & 18.5  
    & 20.0M & 60.0G \\
    \ding{51} & \ding{55} & \ding{55}   & 27.8 & 46.1 & 19.5   & 16.1M   & 53.3G \\
    \ding{55} & \ding{51} & \ding{55}  & 28.2 & 46.7 & 19.8 & 20.2M & 61.4G \\
    \ding{55} & \ding{55} & \ding{51}  & 27.6 & 45.4 & 19.3 & 20.7M & 61.2G \\
    \midrule
    \ding{51} & \ding{51} & \ding{55}  & 28.9 & 47.6 & 20.3  & 16.6M & 58.7G \\
    
    
    \ding{51} & \ding{51} & \ding{51}  
    & \textbf{\textcolor{red!70!black}{29.4}} 
    & \textbf{\textcolor{red!70!black}{48.4}}  
    & \textbf{\textcolor{red!70!black}{21.4}} 
    & 17.4M & 62.9G \\
    \bottomrule
    \end{tabular}
\end{table}

\begin{table}[!t]
\centering
\small
\caption{Ablation studies on the HazyDet and SIMD datasets.}
\label{table4_HazyDet_and_SIMD_Ablation}
\setlength{\tabcolsep}{1.0mm}
\begin{tabular}{ccc|ccc|ccc}
\toprule
\multirow{2}{*}{\textbf{WeKat}} & \multirow{2}{*}{\textbf{MDFC}} & \multirow{2}{*}{\textbf{CPF}}
& \multicolumn{3}{c|}{\textbf{HazyDet}}
& \multicolumn{3}{c}{\textbf{SIMD}} \\
\cmidrule(lr){4-6} \cmidrule(lr){7-9}
& &
& \textbf{AP} & \textbf{AP$_{50}$} & \textbf{AP$_{75}$}
& \textbf{AP} & \textbf{AP$_{50}$} & \textbf{AP$_{75}$} \\
\midrule

\ding{55} & \ding{55} & \ding{55}
& 53.3 & 73.6 & 62.2    
& 63.7 & 78.6 & 72.4 \\

\ding{51} & \ding{55} & \ding{55}
& 54.0 & 73.8 & 62.7
& 63.9 & 78.5 & 73.8 \\

\ding{55} & \ding{51} & \ding{55}
& 54.7 & 73.7 & 63.1
& 64.6 & 79.2 & 74.3 \\

\ding{55} & \ding{55} & \ding{51}
& 54.3 & 73.6 & 62.9
& 65.2 & 80.1 & 74.2 \\
\midrule

\ding{51} & \ding{51} & \ding{55}
& 54.8 & 74.2 & 63.5
& 64.4 & 78.7 & 75.3 \\

\ding{51} & \ding{51} & \ding{51}
& \textbf{\textcolor{red!70!black}{55.0}}
& \textbf{\textcolor{red!70!black}{74.5}}
& \textbf{\textcolor{red!70!black}{63.6}}
& \textbf{\textcolor{red!70!black}{66.6}}
& \textbf{\textcolor{red!70!black}{80.9}}
& \textbf{\textcolor{red!70!black}{77.8}} \\
\bottomrule
\end{tabular}
\end{table}
\subsubsection{Analysis of WeKat Backbone Design}
To further validate the internal design of WeKat, we conduct four complementary analyses covering its principal components, shallow spectral modeling strategy, HSG channel allocation, and optimization behavior.

First, we disentangle the contributions of HSG-WAVE, asymmetric attention, and Group KAN. As shown in Table~\ref{tab:wekat_component_ablation}, HSG-WAVE improves AP while reducing FLOPs. Simply adding standard attention with an MLP does not provide further gains, whereas asymmetric attention increases AP to 27.5\%. Replacing the MLP with Group KAN further improves AP to 27.8\% and AP$_{75}$ to 28.2\%. These results demonstrate the complementary effectiveness of the proposed components rather than a simple accumulation of operators.

Second, we examine the necessity of the wavelet-based shallow modeling in HSG-WAVE. Specifically, we keep the overall HSG routing structure and the remaining backbone unchanged, but replace the wavelet-based spectral decomposition in HSG-WAVE with the frequency-domain design adopted in MDFC. As shown in Table~\ref{tab:wekat_module_ablation}, the original HSG-WAVE achieves better AP/AP$_{50}$/AP$_{75}$ with fewer parameters and lower FLOPs. This is because wavelet decomposition preserves spatial locality while explicitly separating low-frequency structures from high-frequency details, enabling recursive low-frequency refinement to capture broader structural context without interfering with local discriminative responses. By contrast, the FFT-based operation in MDFC emphasizes global spectral interaction, which is more appropriate for cross-scale feature refinement than for shallow structure-aware modeling.

Third, we analyze the sensitivity of HSG with respect to the initial split ratio \(\alpha\) and the Com-Stream allocation ratio \(e\). Here, \(\alpha\) controls the first channel allocation in HSG, while \(e\) determines the relative capacity assigned to the internal Com-Stream. As shown in Table~\ref{tab:hsg_sensitivity}, increasing \(\alpha\) from 0.25 to 0.50 brings a clear improvement, indicating that an appropriate channel allocation is important for balancing feature preservation and transformation. Although \(\alpha=0.75\) achieves slightly higher accuracy, it also increases the computational cost from 53.3G to 69.9G. Therefore, we adopt \(\alpha=0.50\) as a better accuracy-complexity trade-off. For the Com-Stream allocation, \(e=1.00\) provides stable overall performance with moderate FLOPs, while further increasing it to 1.25 brings limited gains.

\begin{figure}[!t]
\centering
\includegraphics[width=.9\columnwidth]{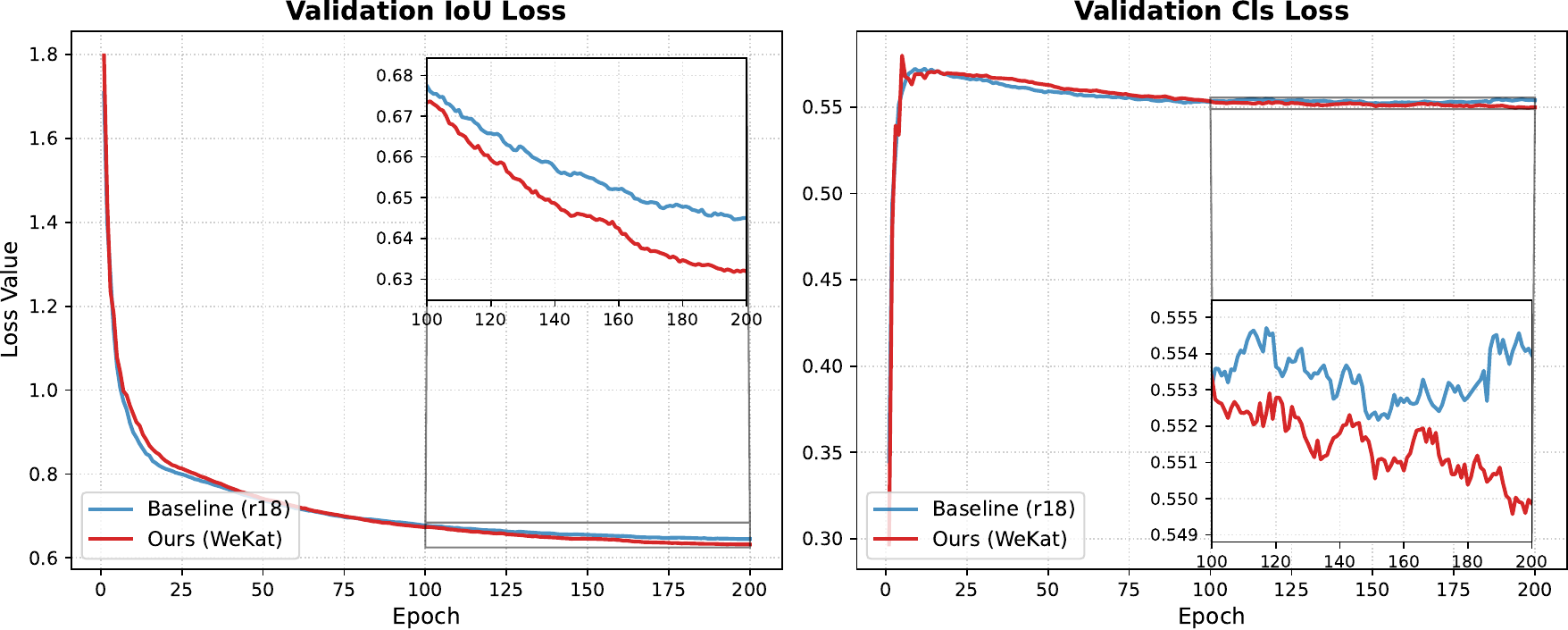} 
\caption{Comparison of validation loss curves between the baseline RT-DETR and the proposed WeKat.}
\label{fig_loss}
\end{figure}

Finally, we further examine the optimization behavior of WeKat by comparing the validation IoU loss and classification loss during training. Fig.~\ref{fig_loss} shows that, relative to the baseline, WeKat consistently attains lower validation IoU and classification losses throughout training, with lower validation losses in the later epochs. In particular, the gap becomes clearer after the initial optimization phase, where WeKat reaches lower final losses for both localization and classification. This suggests that the proposed HSG routing improves feature transformation and information flow, thereby making the backbone easier to optimize.
\begin{table}[!t]
    \centering
    \small
    \caption{Progressive ablation of WeKat on the VisDrone validation set. Asym. and GKAN denote asymmetric attention and Group KAN, respectively. A \ding{55} indicates that the corresponding HSG module is replaced with a conventional C2f \cite{yolov8_ultralytics} feature-extraction block.}
    \label{tab:wekat_component_ablation}
    \setlength{\tabcolsep}{0.5mm}
    \begin{tabular}{cc|ccccc}
    \toprule
    \textbf{HSG-WAVE}
    & \textbf{HSG-AKAT}
    & \textbf{AP}
    & \textbf{AP$_{50}$}
    & \textbf{AP$_{75}$}
    & \textbf{Params}
    & \textbf{GFLOPs} \\
    \midrule

    \ding{51}
    & Asym.+GKAN
    & \textbf{\textcolor{red!70!black}{27.8}}
    & 46.1
    & \textbf{\textcolor{red!70!black}{28.2}}
    & \textbf{\textcolor{red!70!black}{16.1}}
    & \textbf{\textcolor{red!70!black}{53.3}} \\

    \ding{51}
    & Asym.+MLP
    & 27.5 & \textbf{\textcolor{red!70!black}{46.2}} & 27.8 & 16.1 & 53.3 \\

    \ding{51}
    & Attn+MLP
    & 27.1 & 46.0 & 27.1 & 16.3 & 53.6 \\

    \ding{51}
    & \ding{55}
    & 27.4 & 45.9 & 27.4 & 16.0 & 53.0 \\

    \ding{55}
    & \ding{55}
    & 27.3 & 46.0 & 27.3 & 16.3 & 59.0 \\

    \bottomrule
    \end{tabular}
\end{table}

\begin{table}[!t]
    \centering
    \small
    \caption{Ablation on the spectral modeling block in WeKat. We replace the wavelet-based branch in HSG-WAVE with the FFT-based spectral modeling used in MDFC.}
    \label{tab:wekat_module_ablation}
    \setlength{\tabcolsep}{1.5mm}
    \begin{tabular}{lccccc}
    \toprule
    \textbf{Spectral block} & \textbf{AP} & \textbf{AP$_{50}$} & \textbf{AP$_{75}$} & \textbf{Param} & \textbf{FLOPs} \\
    \midrule
    WAVE (Ours) & \textbf{27.8} & \textbf{46.1} & \textbf{28.2} & \textbf{16.1M} & \textbf{53.3G} \\
    MDFC        & 27.7          & 45.9          & 27.9          & 16.3M          & 55.3G \\
    \bottomrule
    \end{tabular}
\end{table}

\begin{table}[!t]
    \centering
    \small
    \caption{Sensitivity analysis of the HSG design in WeKat. 
    \(\alpha\) denotes the initial split ratio, and \(e\) denotes the Com-Stream allocation ratio. 
    \(\dagger\) indicates the default setting.}
    \label{tab:hsg_sensitivity}
    \setlength{\tabcolsep}{1.5mm}
    \renewcommand{\arraystretch}{1.08}
    \begin{tabular}{cc|cccc|c}
    \toprule
    \textbf{\(\alpha\)} & \textbf{\(e\)} 
    & \textbf{AP} & \textbf{AP$_{50}$} & \textbf{AP$_{75}$} & \textbf{AP$_S$} 
    & \textbf{FLOPs} \\
    \midrule
    \multicolumn{7}{c}{\textit{Varying \(\alpha\) with fixed \(e=1.0\)}} \\
    \midrule
    0.25 & 1.00 & 26.7 & 44.6 & 26.9 & 19.0 & 42.0G \\
    \(0.50^{\dagger}\) & \(1.00^{\dagger}\) & 27.8 & 46.1 & 28.2 & 19.5 & 53.3G \\
    0.75 & 1.00 & \textbf{28.2} & \textbf{46.8} & \textbf{28.5} & \textbf{20.1} & 69.9G \\
    \midrule
    \multicolumn{7}{c}{\textit{Varying \(e\) with fixed \(\alpha=0.50\)}} \\
    \midrule
    0.50 & 0.75 & 27.6 & 46.2 & 28.0 & 19.6 & 51.9G \\
    \(0.50^{\dagger}\) & \(1.00^{\dagger}\) & \textbf{27.8} & 46.1 & \textbf{28.2} & 19.5 & 53.3G \\
    0.50 & 1.25 & \textbf{27.8} & \textbf{46.4} & 27.8 & \textbf{19.7} & 55.2G \\
    \bottomrule
    \end{tabular}
\end{table}

\begin{table}[!t]
    \centering
    \small
    \caption{Fine-grained ablation of the spatial, edge-aware, and frequency branches in the multi-domain refinement stage of MDFC on the VisDrone dataset.}
    \label{tab:mdfc_branch_ablation}
    \setlength{\tabcolsep}{1.2mm}
    \begin{tabular}{ccc|ccc|ccc}
    \toprule
    \textbf{Spa.} & \textbf{Edge.} & \textbf{Freq.} 
    & \textbf{AP} & \textbf{AP$_{50}$}   & \textbf{AP$_{75}$}  
    & \textbf{AP$_{S}$} & \textbf{AP$_{M}$}  & \textbf{AP$_{L}$} \\
    \midrule
    \ding{51} & \ding{55} & \ding{55}   & 27.2 & 45.9 & 27.2  & 19.2 & 37.1  & 41.6\\
    \ding{51} & \ding{51} & \ding{55}   & 28.0 & 46.3 & 28.5  & 19.7 & 38.4  & 42.6 \\
    \ding{51} & \ding{55} & \ding{51}   & 28.0 & 46.2 & 28.5  & 19.4 & 38.7  & 42.6 \\
    \ding{55} & \ding{51} & \ding{51}   & 27.7 & 46.2 & 28.1  & 19.5 & 38.1  & 43.3 \\ 
    
    \ding{51} & \ding{51} & \ding{51}  
    & \textbf{\textcolor{red!70!black}{28.2}} 
    & \textbf{\textcolor{red!70!black}{46.7}}  
    & \textbf{\textcolor{red!70!black}{28.8}} 
    & \textbf{\textcolor{red!70!black}{19.8}} 
    & \textbf{\textcolor{red!70!black}{38.8}}  
    & \textbf{\textcolor{red!70!black}{44.0}}  \\
    \bottomrule
    \end{tabular}
\end{table}
\subsubsection{Analysis of MDFC Multi-Domain Design}
To investigate the contribution of each domain-specific branch in MDFC, we independently ablate the spatial, frequency, and edge-aware branches while keeping the remaining configuration unchanged. As shown in Table~\ref{tab:mdfc_branch_ablation}, the spatial branch alone achieves 27.2\% AP. Introducing either the edge-aware or frequency branch increases AP to 28.0\%, indicating that both structural boundaries and global spectral information provide complementary cues to local spatial modeling. Combining the frequency and edge branches without the spatial branch achieves 27.7\% AP, further suggesting that local spatial representation remains an important basis for multi-domain coordination. Integrating all three branches achieves the best overall performance, reaching 28.2\% AP, 28.8\% AP$_{75}$, and 19.8\% AP$_S$. These results verify that the three branches provide complementary information.

\subsubsection{Effectiveness of the Detection-Layer Configuration}
\begin{figure}[!t]
\centering
\includegraphics[width=.9\columnwidth]{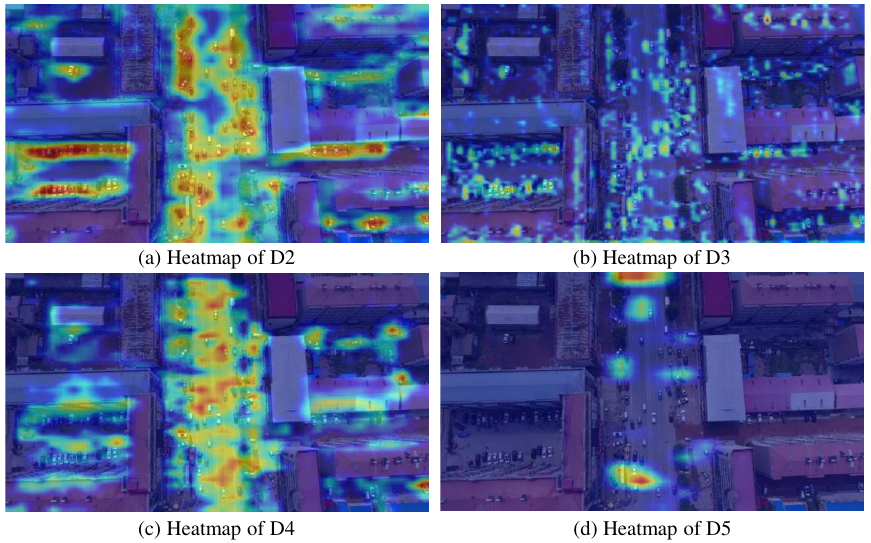} 
\caption{Visualization of feature activations across detection layers for small objects. Here, \( [D2-D5] \) denote detection layers at different feature scales, where the spatial resolution decreases progressively by a factor of 2 at each stage.}
\label{fig4-heatmap-d3d4d5}
\end{figure}

Table~\ref{table5_detectlayer} evaluates how removing the \(S5\) backbone stage and reconfiguring the detection layers affect detection performance. FMC-DETR-B retains the \(S5\) stage and uses three detection layers, \([D3,D4,D5]\), whereas FMC-DETR-T removes \(S5\) and reconfigures the detection pyramid to \([D2,D4]\). This design shifts the detection scales toward higher-resolution features while reducing the number of detection layers from three to two. As shown in Table~\ref{table5_detectlayer}, we first examine the effect of shifting the detection pyramid toward higher-resolution features under the WeKat-only configuration. As the lowest-resolution backbone stage, \(S5\) provides high-level semantic information, but repeated downsampling may weaken the fine-grained spatial cues of small objects. We therefore remove the final downsampling stage, such that the backbone terminates at \(S4\) and the corresponding \(D5\) detection feature is no longer generated. To maintain three detection scales in this comparison, \(S2\), which is not used in the original detection pyramid, is incorporated into cross-scale feature fusion to generate \(D2\). The detection pyramid is consequently shifted from \([D3,D4,D5]\) to \([D2,D3,D4]\). This modification improves AP from 27.8\% to 30.9\% and AP$_S$ from 19.5\% to 23.5\%, corresponding to gains of 3.1 and 4.0 percentage points, respectively. As shown by the heatmap comparison in Fig.~\ref{fig4-heatmap-d3d4d5}, \(D2\) and \(D4\) exhibit stronger and more spatially concentrated responses to small objects than \(D3\) and \(D5\). We then examine whether all three shifted detection scales are necessary. The compact \([D2,D4]\) configuration achieves 31.6\% AP, 50.9\% AP$_{50}$, and 23.9\% AP$_S$, outperforming both \([D2,D3,D4]\) and \([D3,D4]\). These results indicate that \(D2\) preserves fine-grained spatial information for small objects, while \(D4\) provides stronger semantic context, forming an effective detail-semantic combination. With MDFC and CPF further integrated, the resulting FMC-DETR-T variant adopts $[D2,D4]$ and achieves 33.7\% AP, 53.6\% AP$_{50}$, and 25.8\% AP$_S$.
\begin{table*}[!t]
    \centering
    \caption{Ablation Study on the Effect of \( S5 \) Backbone Layer and Detection Head Design. 
    ``\ding{51}'' and ``\ding{55}'' indicate the presence or absence of the \( S5 \) stage, respectively.}
    \label{table5_detectlayer}
    \setlength{\tabcolsep}{4mm}
    \begin{tabular}{c|c|c|ccc|c|cc}
    \toprule
    \textbf{Model} & \textbf{S5} &  \textbf{Detect Layer}  & \textbf{AP} & \textbf{AP$_{50}$} & \textbf{AP$_{75}$} & \textbf{AP$_{S}$}  & \textbf{Param} & \textbf{FLOPs} \\
    \hline
    Baseline  & \ding{51} & $[D3,D4,D5]$ & 26.7 & 44.6 & 26.9 & 18.5 & 20.0M & 60.0G\\
    WeKat     & \ding{51} & $[D3,D4,D5]$ & 27.8 & 46.1 & 28.2 & 19.5 & 16.1M & 53.3G\\
    \hline
    WeKat     & \ding{55} & $[D2,D3,D4]$ & 30.9 & 50.2 & 31.8 & 23.5 & 12.9M & 120.1G\\
    WeKat     & \ding{55} & $[D3,D4]$  & 30.8 & 50.2 & 31.5 & 23.3 & 12.8M & 103.0G\\
    WeKat     & \ding{55} &  $[D2,D4]$   & 31.6 & 50.9 & 32.9 & 23.9 & 12.8M & 115.8G\\
    \hline
    FMC-DETR-B & \ding{51} & $[D3,D4,D5]$ & 29.4 & 48.4 & 30.2 & 21.4 & 17.4M & 62.9G\\
    FMC-DETR-T & \ding{55} & $[D2,D4]$     & 33.7 & 53.6 & 35.4 & 25.8 & 13.8M & 146.8G\\
    \bottomrule
    \end{tabular}
\end{table*}

\subsubsection{Activation-function Comparison}
The proposed FMC-DETR adopts SiLU as the activation function, following the common practice in recent mainstream object detectors, including YOLO-series and DETR-style models. Compared with ReLU, SiLU provides a smoother nonlinear mapping and avoids the hard truncation of negative responses. This property helps preserve weak feature activations and facilitates gradient propagation, which is beneficial for aerial-view small-object detection where discriminative target cues are often subtle and easily suppressed. To isolate the effect of the activation function, we conduct this comparison using the WeKat-only configuration in Table~\ref{table4_visdrone_ablation}. Specifically, SiLU is replaced with ReLU while all other settings are kept unchanged. As shown in Table~\ref{tab_activation_function}, SiLU provides more favorable feature representations for the WeKat backbone, supporting its use in the proposed framework.

\subsubsection{PR-curve Analysis}
\begin{figure}[!t]
\centering
\includegraphics[width=.9\columnwidth]{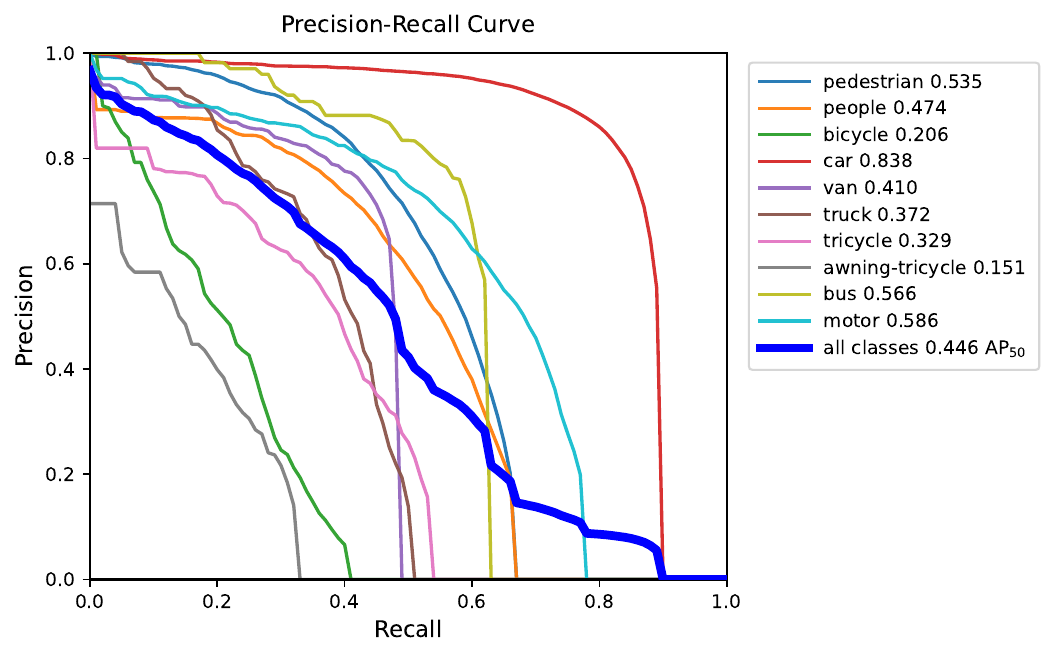}
\caption{Precision--Recall curves of the baseline RT-DETR on the VisDrone dataset at IoU=0.50.}
\label{fig:pr_rtdetr}
\end{figure}

\begin{figure}[!t]
\centering
\includegraphics[width=.9\columnwidth]{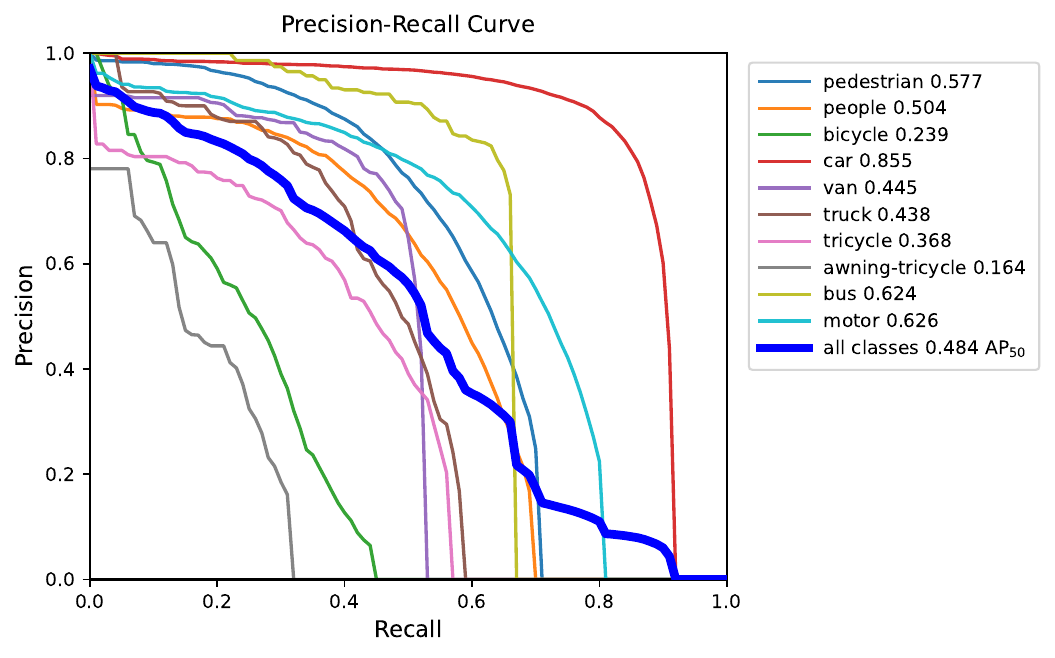}
\caption{Precision--Recall curves of the proposed FMC-DETR-B on the VisDrone dataset at IoU=0.50.}
\label{fig:pr_fmcdetr}
\end{figure}

In addition to the qualitative visualization of feature responses and detection results, we further provide precision-recall (PR) curve visualizations to analyze the detection behavior from the perspective of confidence-threshold variation. In this work, we follow the COCO evaluation protocol, where the main metrics include AP, AP$_{50}$, and AP$_{75}$. Under this protocol, AP$_{50}$ is computed as the area under the precision-recall curve at a fixed IoU threshold of 0.50, where the curve is obtained by varying the confidence threshold. Fig.~\ref{fig:pr_rtdetr} and Fig.~\ref{fig:pr_fmcdetr} show the PR curves of the baseline RT-DETR and the proposed FMC-DETR-B on the VisDrone dataset, respectively. Overall, FMC-DETR-B achieves a more favorable precision-recall trade-off, with the all-class AP$_{50}$ increasing from 44.6\% to 48.4\%. Improvements can be observed in most categories, suggesting that the proposed method improves detection robustness for small objects under cluttered aerial-view conditions.

\begin{table}[t!]
\caption{Comparison of activation functions on the VisDrone dataset.}
\label{tab_activation_function}
\centering
\small
\setlength{\tabcolsep}{2.5mm}
\begin{tabular}{ccccc}
\midrule
\textbf{Activation Function} & \textbf{AP} & \textbf{AP$_{75}$} & \textbf{AP$_{50}$} & \textbf{AP$_{S}$} \\
\midrule
ReLU                         & 27.3 & 27.6 & 45.2 & 19.4 \\
SiLU                         & 27.8 & 28.2 & 46.1 & 19.5 \\
\bottomrule
\end{tabular}
\end{table}

\begin{table}[t]
\centering
\caption{Inference efficiency under batch size 1. Latency and FPS are averaged over five runs on an NVIDIA RTX 4090D GPU with an input size of $640 \times 640$.}
\label{tab:inference_efficiency}
\begin{tabular}{lccc}
\midrule
Method & Model size & Latency (ms/img) & FPS (img/s) \\
\midrule
RT-DETR-R18  & 77.0 MB & 15.4 & 64.8 \\
FMC-DETR-B  & 67.1 MB & 22.7 & 44.0 \\
FMC-DETR-T  & 53.4 MB & 20.7 & 48.4 \\
\bottomrule
\end{tabular}
\end{table}
\subsection{Visualization Analysis}
\begin{figure}[t]
\centering
\includegraphics[width=\columnwidth]{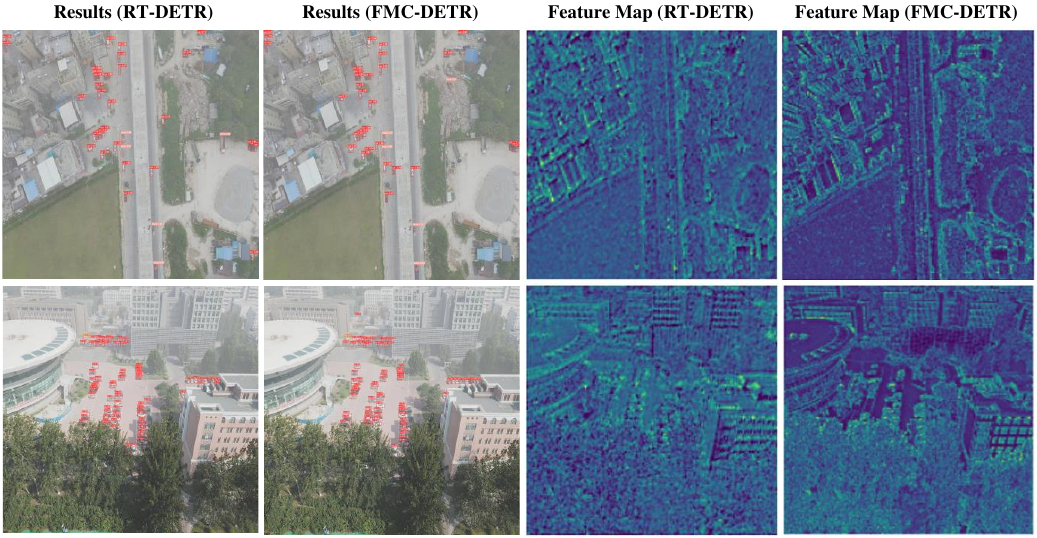} 
\caption{Visualizations of the detection results of baseline and our proposed FMC-DETR-B on HazyDet.}
\label{fig8-heatmap-hazydet}
\end{figure}

\begin{figure}[t]
\centering
\includegraphics[width=\columnwidth]{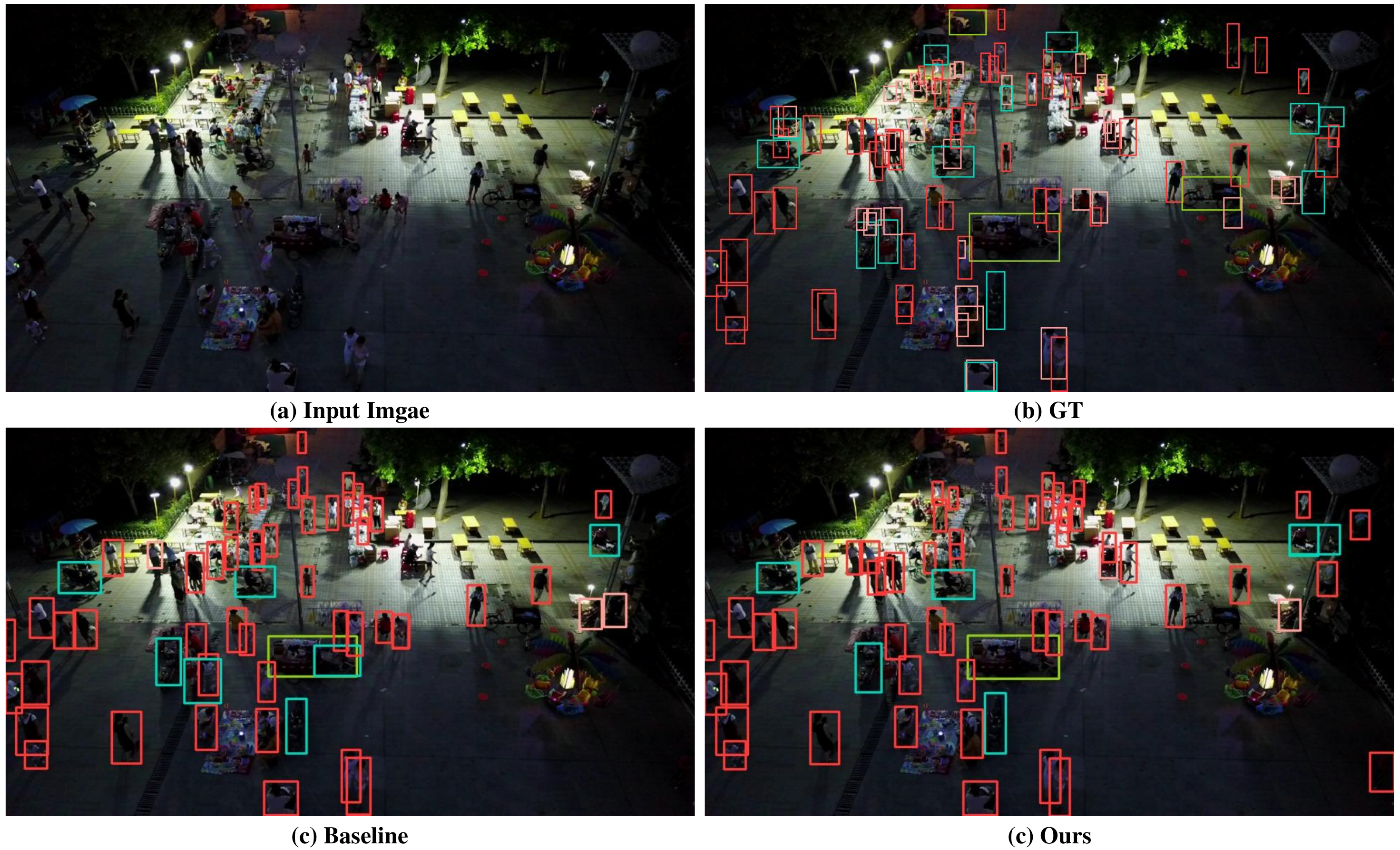} 
\caption{Representative failure cases on the VisDrone validation set.}
\label{fig9}
\end{figure}

\begin{figure*}[t]
\centering
\includegraphics[width=.7\textwidth]{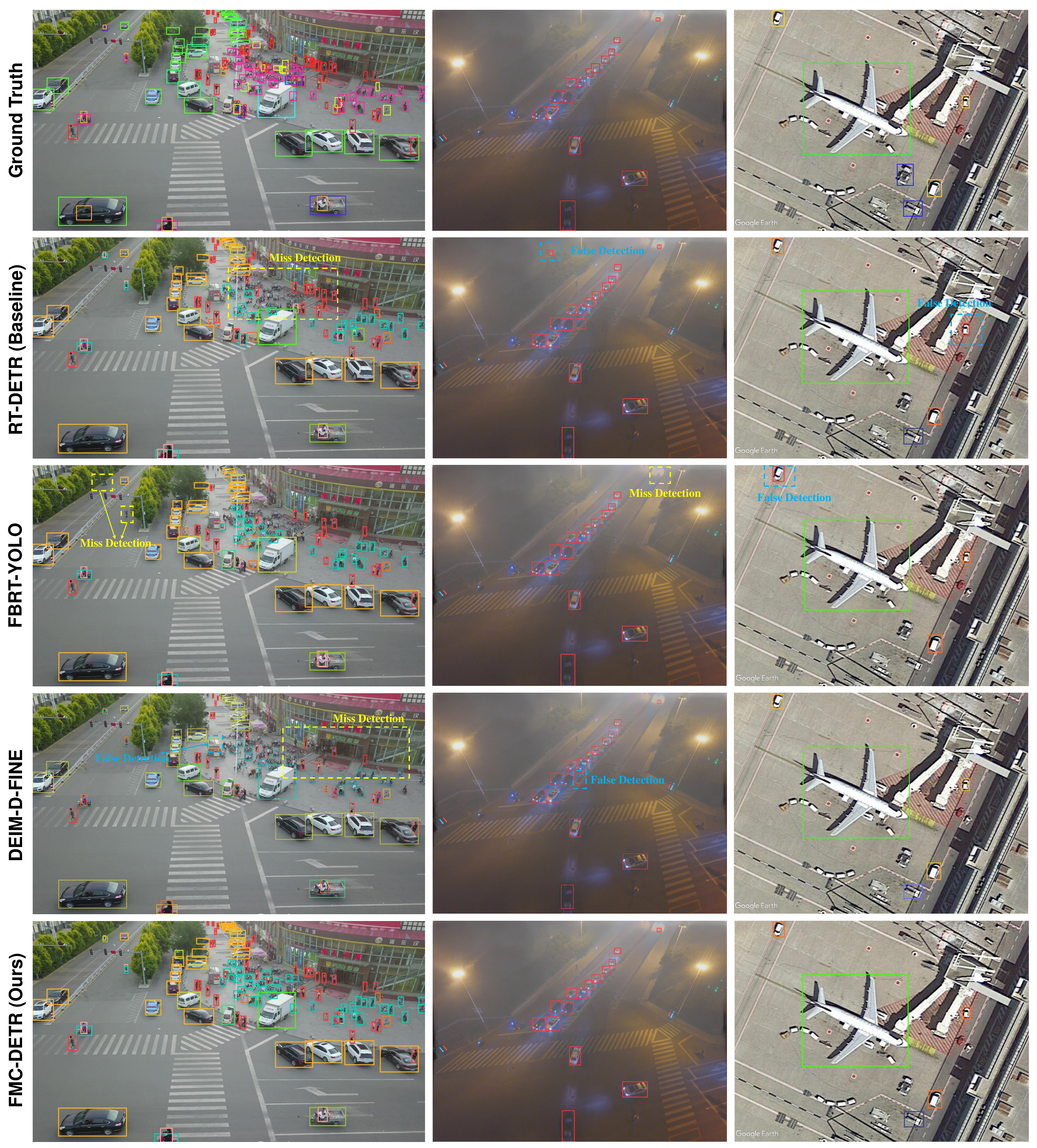} 
\caption{Qualitative comparison of detection results produced by different methods across three aerial-view datasets. Each column corresponds to a different dataset, and the last row shows the results of the proposed framework, which exhibits more accurate localization and fewer missed or false detections.}
\label{fig5_visdrone}
\end{figure*}

To further demonstrate the effectiveness of the proposed framework in aerial-view scenarios, we visualize both feature responses and detection results in Figs.~\ref{fig8-heatmap-hazydet}--\ref{fig5_visdrone}. Compared with the baseline, FMC-DETR-B produces clearer structural representations and more informative contextual responses, enabling more accurate object localization. It better captures both holistic object shapes and fine-grained boundary details, resulting in more precise predictions. Furthermore, the qualitative comparisons show that FMC-DETR-B achieves stronger multi-scale detection performance, particularly for small objects. Relative to RT-DETR, our method localizes small targets more accurately and pays greater attention to their surrounding contextual regions, indicating a more effective use of contextual cues. We further present representative failure cases in Fig.~\ref{fig9}. Although FMC-DETR-B recovers more small and partially occluded objects than the baseline, residual errors remain when objects occupy only a few pixels or suffer from severe occlusion, motion blur, and low illumination. Visually similar categories may also cause occasional classification ambiguity. Overall, these observations highlight the robustness of the proposed framework for multi-object detection in complex aerial scenes while also identifying potential directions for improving robustness under extreme imaging conditions.

\subsection{Limitations and Discussion}
Despite its favorable detection accuracy, the proposed method still has several limitations. Although our design reduces the number of parameters, it does not achieve a proportional reduction in computational cost. The multi-branch feature coordination and compact feature fusion modules introduce additional operations, leading to relatively higher FLOPs and potentially slower practical inference. As reported in Table~\ref{tab:inference_efficiency}, FMC-DETR-B achieves 44.0 FPS with a latency of 22.7 ms per image and a model size of 67.1 MB. FMC-DETR-B is slower than the original RT-DETR-R18 baseline because it intentionally incorporates stronger feature representation modules to improve remote-sensing object detection in challenging scenes. In particular, the KAN-based component introduces more complex nonlinear modeling than standard convolutional operations, while current KAN-style operators remain less optimized at the kernel and inference-engine levels, leading to higher computational cost and GPU inference overhead. FMC-DETR-T reconfigures the detection head from \([D3,D4,D5]\) to \([D2,D4]\), incorporating the higher-resolution \(D2\) feature to enhance small-object detection while reducing the total number of detection layers from three to two. Notably, this configuration achieves slightly better practical inference efficiency than FMC-DETR-B. We attribute this practical speedup mainly to the removal of the KAN-based $S5$ stage and the use of only two detection layers, which reduce the parameter count and remove operations that are less efficiently supported by current GPU kernels. This result suggests that carefully pruning high-cost representation stages while preserving small-object-sensitive detection pathways can provide a better balance between efficiency and detection capability. This observation is encouraging and deserves further exploration in future lightweight aerial object detection models. In future work, we will improve the practical efficiency of FMC-DETR-B through lightweight architectural redesign, operator re-parameterization, and deployment-oriented optimization.
\section{Conclusion}
In this paper, we propose FMC-DETR, a remote sensing object detector built upon frequency-aware modeling and multi-domain feature coordination. Specifically, we first design the WeKat backbone to enhance shallow structural representations and strengthen deep nonlinear feature modeling. Then, we introduce the MDFC module to coordinate spatial, frequency, and structural cues, enabling the network to refine multi-domain feature correlations and extract more discriminative small-object representations. Finally, we develop the CPF module to progressively refine informative spatial responses and aggregate diverse multi-branch features, enhancing local contextual modeling while maintaining stable feature propagation. Extensive qualitative and quantitative experiments validate the effectiveness of the proposed framework, demonstrating competitive performance without relying on any pretrained weights.



 
%

\bibliography{References}
\bibliographystyle{ieeetr}

\end{document}